\documentclass[10pt,twocolumn,letterpaper]{article}

\usepackage{cvpr}              %

\usepackage{multirow}
\usepackage{enumitem}
\usepackage{rotating}
\usepackage{makecell}
\usepackage{marvosym}
\usepackage{pifont}
\usepackage{ragged2e}
\usepackage{caption}
\usepackage{subcaption}
\usepackage[normalem]{ulem}
\usepackage[pagebackref,breaklinks,colorlinks,allcolors=cvprblue]{hyperref}
\usepackage[capitalize]{cleveref}
\crefname{section}{Sec.}{Secs.}
\Crefname{section}{Section}{Sections}
\Crefname{table}{Table}{Tables}
\crefname{table}{Tab.}{Tabs.}
\usepackage{color}
\usepackage{xcolor}
\usepackage{color, colortbl}
\usepackage{booktabs}
\usepackage{float}
\usepackage{wrapfig}
\usepackage{lineno}
\usepackage{colortbl}
\usepackage{amsmath,amssymb,amsfonts}
\usepackage{algorithmic}
\usepackage{graphicx}
\usepackage{float}

\newcommand{\datasetfullname}{Large Generated Human Multi-View Dataset}
\def\datasetname{\emph{HuGe100K}\xspace}
\def\modelname{\emph{IDOL}\xspace}
\definecolor{myPurple}{rgb}{0.4, .0, .8}
\definecolor{myGreen}{rgb}{0, .8, .3}
\definecolor{myRed}{rgb}{0.8, .2, .2}
\definecolor{myOrange}{rgb}{0.8, 0.45, 0.0}
\definecolor{myBlue}{rgb}{.0, .0, 1.0}

\definecolor{cvprblue}{rgb}{0.21,0.49,0.74}

\title{IDOL: Instant Photorealistic 3D Human Creation from a Single Image}

\author{Yiyu Zhuang$^{1, 4*}$ \quad Jiaxi Lv$^{2, 4*}$ \quad Hao Wen$^{3, 4*}$ \quad Qing Shuai$^{4}$ \quad Ailing Zeng$^{4\dagger}$ \quad Hao Zhu$^{1\dagger}$ \\
Shifeng Chen$^{2,5}$ \quad Yujiu Yang$^{3}$ \quad Xun Cao$^{1}$ \quad Wei Liu$^{4}$ \\
 {\footnotesize  
$^1$ Nanjing University, $^2$ Shenzhen Institute of Advanced Technology, Chinese Academy of Sciences}\\ 
{\footnotesize  
$^3$ Tsinghua University, $^4$ Tencent, $^5$ Shenzhen University of Advanced Technology}
}

\begin{document}

\twocolumn[{%
\renewcommand\twocolumn[1][]{#1}%
\maketitle
\vspace{-0.5in}
\begin{center}
    \href{https://yiyuzhuang.github.io/IDOL/}{\textbf{ https://yiyuzhuang.github.io/IDOL/}}
\end{center}

\vspace{-0.25in}
\begin{center}
    \centering
    \includegraphics[width=0.95\linewidth]{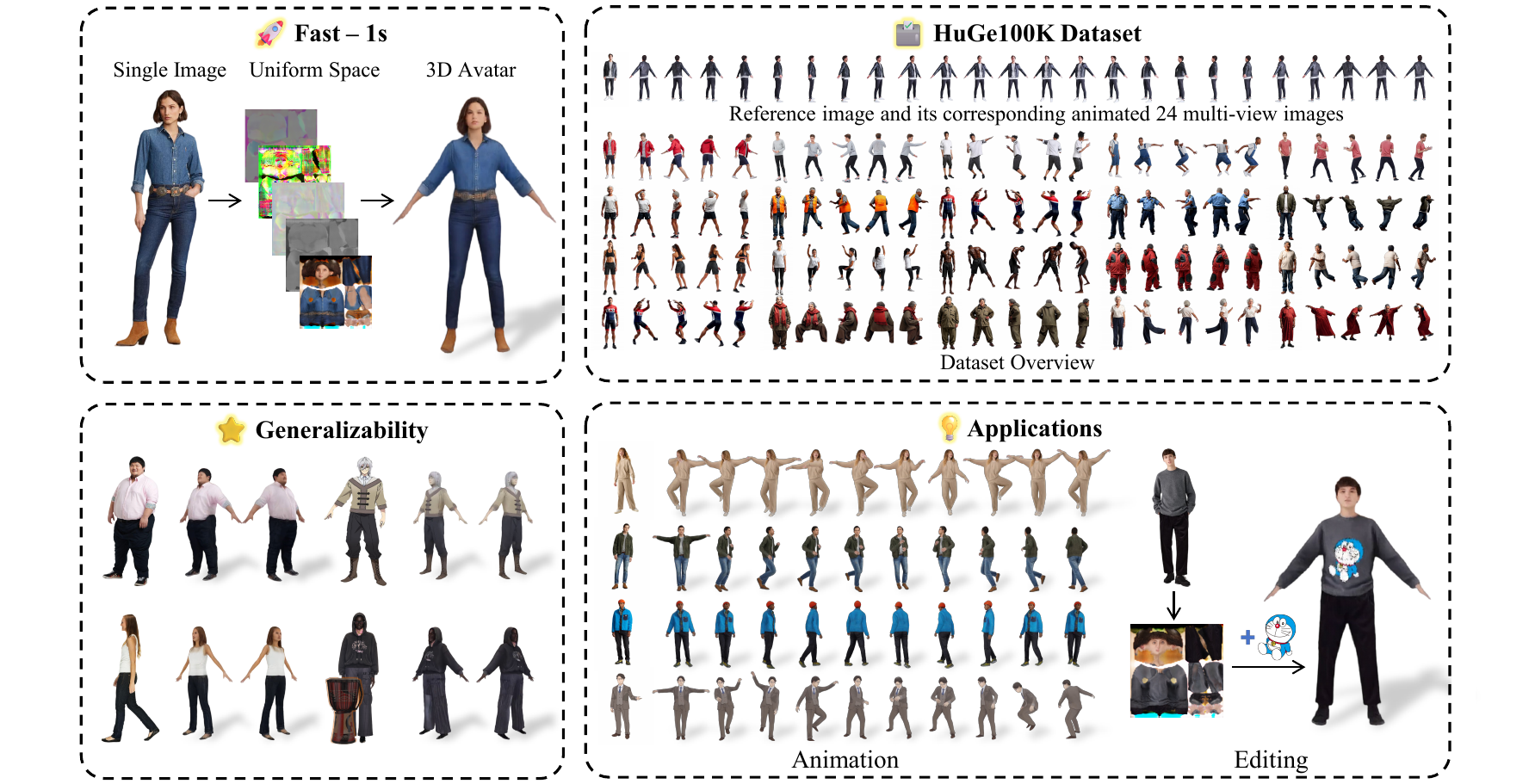}
    \vspace{-0.12in}
    \captionof{figure}{
    This work introduces (a) IDOL, a feed-forward, single-image human reconstruction framework that is fast, high-fidelity, and generalizable; (b) Utilizing the proposed \datasetfullname{} consisting of $100K$ multi-view subjects, our method exhibits exceptional generalizability in handling diverse human shapes, cross-domain data, severe viewpoints, and occlusions; (c) With a uniform structured representation, the avatars can be directly animatable and easily editable.
    }
\label{fig:title}
\end{center}
}]

\begingroup
\renewcommand{\thefootnote}{}
\footnotetext{$^{}$ Work done during the internship at Tencent by Yiyu Zhuang, Jiaxi Lv, and Wenhao.}
\footnotetext{$^{*}$Equal contributions}
\footnotetext{$^{\dagger}$Corresponding authors}
\endgroup

\begin{abstract}

Creating a high-fidelity, animatable 3D full-body avatar from a single image is a challenging task due to the diverse appearance and poses of humans and the limited availability of high-quality training data. To achieve fast and high-quality human reconstruction, this work rethinks the task from the perspectives of dataset, model, and representation.
First, we introduce a large-scale HUman-centric GEnerated dataset, HuGe100K, consisting of 100K diverse, photorealistic sets of human images. Each set contains 24-view frames in specific human poses, generated using a pose-controllable image-to-multi-view model.
Next, leveraging the diversity in views, poses, and appearances within HuGe100K, we develop a scalable feed-forward transformer model to predict a 3D human Gaussian representation in a uniform space from a given human image. 
This model is trained to disentangle human pose, body shape, clothing geometry, and texture. 
The estimated Gaussians can be animated without post-processing. 
We conduct comprehensive experiments to validate the effectiveness of the proposed dataset and method. Our model demonstrates the ability to efficiently reconstruct photorealistic humans at 1K resolution from a single input image using a single GPU instantly. Additionally, it seamlessly supports various applications, as well as shape and texture editing tasks.

\end{abstract}
    
\section{Introduction}

Reconstructing 3D clothed avatars from a single image is crucial in user-friendly virtual reality, gaming, and 3D content creation. This task involves mapping 2D images to 3D models, a highly ill-posed and challenging problem due to  the complexity of clothing and the diversity of human poses.
Learning-based methods trained on publicly available 3D human models~\cite{zheng2019deephuman, han2023high, xiong2024mvhumannet, twindom, renderpeople} have improved reconstruction quality, but their generalization and quality remain limited~\cite{saito2019pifu, saito2020pifuhd, zheng2021pamir, xiu2022icon, xiu2022econ, zhang2024sifu, zhang2024humanref}. 
Techniques that integrate parametric body estimation~\cite{huang2020arch, zheng2021pamir, xiu2022icon}, loop optimization~\cite{xiu2022icon, xiu2022econ, zhang2024sifu}, and diffusion model priors~\cite{li2024pshuman, HumanLRM2024, HumanSGD:2023} enhance performance but often result in slower and fragile reconstructions.
Recently, large generic image-to-3D reconstruction and generation models~\cite{voleti2025sv3d,hong2023lrm,tang2025lgm,xu2024instantmesh,wu2024unique3d,li2023instant3d}, leveraging large-scale datasets~\cite{deitke2023objaverse,deitke2024objaverse-xl,luo2024cap3d} or pre-trained diffusion models~\cite{podell2023sdxl,rombach2022highresolutionimagesynthesislatent,blattmann2023svd}, boost capabilities but still struggle with real-life human reconstruction due to the scarcity of photorealistic 3D human data.

Back to the first principle, we rethink existing data, models, and representations. \textbf{First}, current data acquisition is limited by the photographing or scanning of individual subjects, making them inherently unscalable. For instance, the largest publicly available human dataset, MVHumanNet~\cite{xiong2024mvhumannet}, contains only thousands of subjects and clothing variations, far from the dataset scale required to achieve robust model generalization across diverse input images.
\textbf{Second}, in the realm of animatable reconstruction models, existing approaches predominantly rely on either reconstruct-and-rig methods that necessitate manual post-processing or entangle human parametric models (\emph{e.g.}, SMPL(-X)~\cite{loper2015smpl,pavlakos2019expressive}) for optimization. However, disentangling human pose and body shape, clothed geometry, and texture could simplify the learning process, avoid error accumulation in parameter estimation, and improve efficiency. 
Furthermore, when focusing solely on human reconstruction, the incorporation of multi-view image generation or refinement models warrants careful consideration~\cite{xu2024instantmesh,peng2024charactergen,HumanLRM2024}. While these models can introduce finer details, they may also lead to inconsistencies across different views.
\textbf{Third}, the community of 3D vision develops various representations tailored to application-specific requirements and technological advancements \cite{kerbl20233d, mildenhall2021nerf}. For human-centric applications, ideal representations should be well-structured and expressive to facilitate easy rigging and editing \cite{li2024animatable, hu2024gaussianavatar}, as well as capable of high-resolution and fast rendering \cite{kerbl20233d} to enhance both functionality and realism.

In this work, we introduce a scalable pipeline for training a simple yet efficient feed-forward model for instant photorealistic human reconstruction. 
We present \datasetname, a large-scale dataset comprising over \textbf{2.4M} high-resolution (\textbf{$896\times640$}) multi-view images of \textbf{100K} diverse subjects. To facilitate comprehensive 3D human reconstruction, we develop a scalable data creation pipeline that integrates synthetic and real-world data, ensuring a wide range of attributes such as age, shape, clothing, race, and gender. 
Building upon this dataset, we introduce a novel feed-forward transformer model \modelname that efficiently predicts 3D human avatars with photorealistic textures and accurate geometry. Our approach leverages a pretrained encoder \cite{khirodkar2025sapiens} and a transformer-based backbone for feature extraction and fusion, enabling instant reconstruction (\textbf{within 1 second} on an A100 GPU) without relying on generative models \cite{zhang2024sifu} for refinement. 
By integrating a uniform representation for the 3D human, our model achieves enhanced texture completion and ensures that the reconstructed humans are naturally animatable.
Extensive evaluations demonstrate that our method excels in diverse and challenging scenarios, offering superior consistency and quality compared to existing techniques. Additionally, the efficient representation of our model enables seamless applications in animation, editing, and other downstream tasks.

Our contributions can be summarized as follows:

\begin{itemize}
    \item  We rethink single-view 3D human reconstruction from the perspectives of data, model, and representation. We introduce a scalable pipeline for training a simple yet efficient feed-forward model, named \modelname, for instant photorealistic human reconstruction.
    
    \item  We develop a data generation pipeline and present \datasetname, a large-scale multi-view human dataset featuring diverse attributes, high-fidelity, high-resolution appearances, and a well-aligned SMPL-X model.
    
    \item Our comprehensive study demonstrates that leveraging large-scale generated training data significantly enhances model performance and generalizability, paving the way for further scaling up 3D human reconstruction models.

\end{itemize}

\section{Related Work}
Recent advancements in 3D avatar reconstruction \cite{gao2022mps, zhuang2022mofanerf, yu2024hifi, he2024magicman, gao2024contex, li2022neurips, huang2024dreamwaltz-g} and representation learning \cite{wu2023high, gao2024mani, zhuang2023neai, zhuang2023anti} have greatly improved human modeling. Most relevant to our work are recent efforts on 3D human datasets and single-image human reconstruction.

\subsection{3D Human Datasets}

High-precision 3D human models typically rely on scanning or multi-view camera acquisition systems. Scan-based datasets, such as THuman2.0~\cite{yu2021function4d}, Twindom~\cite{twindom}, and 2K2K~\cite{han2023high}, provide high-fidelity 3D geometries but are limited by a small number of subjects, simple standing poses, and non-human artifacts (\emph{e.g.}, 2K2K). In contrast, multi-view acquisition systems~\cite{peng2021neural,liu2021neural,yu2020humbi,li2021ai,cai2022humman,cheng2022generalizable,isik2023humanrf,cheng2023dna,xiong2024mvhumannet, he2024head360, zhuang2024towards} facilitate the collection of larger datasets with more flexible actions. However, these datasets often face challenges such as biased indoor lighting, fixed viewpoints, and limited scale. Large-scale synthetic and real datasets for generic objects, including Objaverse and MVImgNet~\cite{deitke2023objaverse,deitke2024objaverse-xl,yu2023mvimgnet}, address open-world reconstruction but lack specificity for human models. To overcome these limitations, the proposed \datasetname significantly scales up dataset size, increasing subject diversity by over 100 times compared to previous datasets. It emphasizes diversity in pose, shape, viewpoint, and clothing, paving the way for large-scale model training. We compare existing datasets with ours in Tab. \ref{tab:datasets}.

\begin{table}[ht]
  \centering
  \resizebox{0.45\textwidth}{!}{
\begin{tabular}{llrrc}
\toprule
\textbf{Type} & \multicolumn{1}{l}{\textbf{Dataset}} & \multicolumn{1}{r}{\textbf{\#Frames}} & \multicolumn{1}{r}{\textbf{IDs}} & \multicolumn{1}{c}{\textbf{SMPL(-X)}}  \\
\midrule
\multirow{4}{*}{\rotatebox{90}{3D Scans}} 
      & THuman \cite{zheng2019deephuman} & -   &200    &\ding{52}    \\
    & THuman2.1 \cite{yu2021function4d}  & -   & 2500    &\ding{52}   \\
      & 2K2K \cite{han2023high}   & - & 2050 &  \ding{56}   \\
      &X-Avatar~\cite{shen2023x} & 35.5K &20&\ding{52}\\
\midrule

\multirow{9}{*}{\rotatebox{90}{Multi-view Images}} 

      & ZJU-MoCap \cite{peng2021neural}&   180K     & 10     &\ding{52}       \\
      & Neural Actor \cite{liu2021neural}&   250K     & 8     &\ding{52}     \\
      & HUMBI \cite{yu2020humbi}&  26M     & 772     &\ding{52}     \\
      & AIST++ \cite{li2021ai}&   10.1M     & 30     &\ding{52}     \\
      & HuMMan \cite{cai2022humman}& 60M   &  1000    & \ding{52}     \\
      & GeneBody \cite{cheng2022generalizable}&  2.95M  &  50     & \ding{52}     \\
      & ActorsHQ \cite{isik2023humanrf}&  40K   &  8    &  \ding{56}   \\
      & DNA-Rendering \cite{cheng2023dna}&  67.5M     &  500    & \ding{52}   \\
      & MVHumanNet \cite{xiong2024mvhumannet}&   \textbf{645.1M}    &  4500    & \ding{52}   \\
      
\midrule

\multirow{1}{*}{Ours} 

      & \textbf{\datasetname{}} & $>$ 2.4M & \textbf{100K}  &\ding{52} \\
    \bottomrule
    
    \end{tabular}
     }
    \caption{Comparisons of related datasets. \emph{\datasetname} is a large-scale generated multi-view human dataset with $100K$ diverse high-fidelity humans.}
\label{tab:datasets}
\end{table}

\subsection{Single-Image Human Reconstruction}

For clothed 3D reconstruction methods, implicit representation methods such as PIFU~\cite{saito2019pifu}, PIFU-HD~\cite{saito2020pifuhd}, ARCH~\cite{huang2020arch, he2021arch++}, and PaMIR~\cite{zheng2021pamir} have been widely adopted. To enhance reconstruction quality and generalizability, loop optimization techniques integrate implicit representations with explicit or hybrid human priors for improved robustness~\cite{xiu2022icon,xiu2022econ,zhang2023globalcorrelated,HumanSGD:2023}. Similarly, GTA \cite{zhang2024global} and SIFU \cite{zhang2024sifu} entangle SMPL priors and side-view conditioned features to enhance feature representation. However, these approaches rely heavily on the accuracy of SMPL estimates and are computationally expensive, typically requiring several minutes for inference. Moreover, the resulting 3D representations are not drivable.
Recent advancements focus on recovering animatable 3D humans from single images~\cite{huang2022elicit,hu2023sherf,corona2023structured}. Some methods benefit from the pre-trained diffusion models or large reconstruction models~\cite{peng2024charactergen,HumanLRM2024}. Nevertheless, all methods are constrained by the limitations of training datasets, resulting in suboptimal texture detail and limited generalizability. The latest work, E3Gen~\cite{zhang20243gen}, combines UV maps and Gaussian splatting~\cite{kerbl20233d} to directly generate Gaussian attribute maps in UV space from images, which does not support arbitrary image inputs.

To generate consistent human images or videos given a single image, recent diffusion video models animate a static image with pose-controllable video conditions~\cite{hu2024animate,xu2024magicanimate,wang2024vividpose,chang2023magicpose,wang2024vividpose,chang2023magicpose,zhu2024champ}. Specifically, Champ~\cite{zhu2024champ} incorporate several rendered maps obtained from SMPL sequences, alongside skeleton-based motion guidance, to enrich the conditions to the latent diffusion model. However, they all fail to generate a 360-degree video and animate precise expressive SMPL-X-based humans due to insufficient training data.

\begin{figure*}
  \centering
  \includegraphics[width=1.0\linewidth]{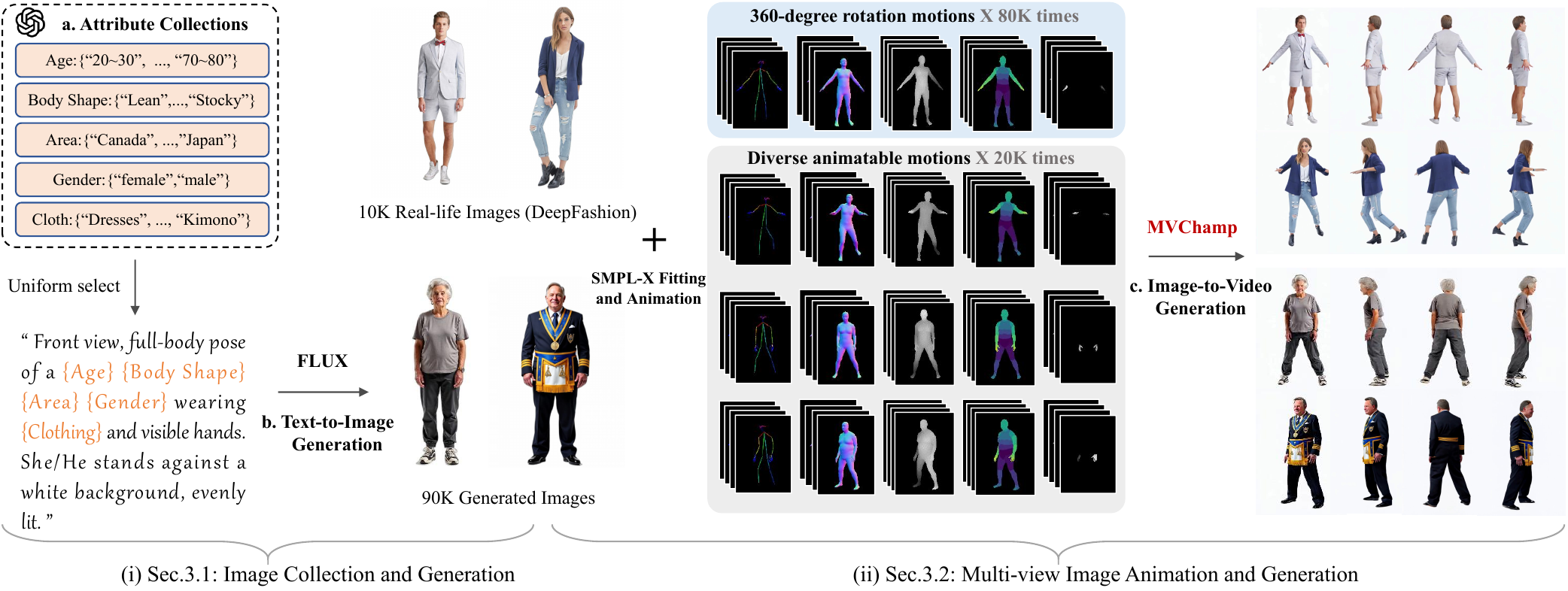}
  \vspace{-20pt}
  \caption{Pipeline for constructing our \datasetname{}. Diverse attribute combinations from GPT-4 templates create text prompts, generating synthetic images via FLUX, combined with real images from DeepFashion. SMPL-X fitting produces multi-view pose sequences with 360-degree rotations and diverse animatable motions. MVChamp then converts these sequences into multi-view images, ensuring 3D consistency in the dataset.}
  \label{fig:dataset_pipeline}
\end{figure*}

\section{Dataset Creation}
\label{sec:dataset}

Drawing inspiration from the latest advancements in generalizable large models, our core insight is to develop a large-scale reconstruction model with exceptional generalization capabilities. The crux of this endeavor lies in creating a comprehensive and high-quality digital human dataset. In this section, we introduce the creation of a large-scale human-centric generated dataset comprising over \textbf{2.4M} high-resolution (896 × 640) multi-view images from more than \textbf{100K} diverse subjects.

As shown in Fig.~\ref{fig:dataset_pipeline}, the data generation pipeline consists of two stages. Firstly, diverse text prompts are generated by large language models incorporating various human-centric attributes, and high-quality images are synthesized using text-to-image generation models (Sec.~\ref{sec: img_gen}). Secondly, we train a multi-view video generation model, MVChamp, conditioned on rendered full-body motions. With this, a large-scale multi-view images dataset is established by feeding both the synthesized and captured images (Sec.~\ref{sec: mvchamp}). Lastly, the statistics and characteristics of the \datasetname are demonstrated in Sec.~\ref{sec: statistic}.
\subsection{Image Collection and Generation}
\label{sec: img_gen}

Given the limited amount of existing datasets~\cite{xiong2024mvhumannet, cai2022humman, peng2021neural, yu2021function4d} and legal concerns regarding portrait and copyright issues, we propose leveraging an image generation model to construct a large-scale, diverse, and high-fidelity dataset of full-body human images. %
Using the latest text-to-image model Flux~\cite{flux}, we design descriptive prompts based on the required attributes for human figures, ensuring diversity across \textit{area}, \textit{clothing}, \textit{body shape}, \textit{age}, and \textit{gender} shown in Fig.~\ref{fig:dataset_pipeline}(i). To avoid long-tail attribute distributions and ensure uniform coverage of traits, we apply uniform sampling in attribute selection for the image generation process. This approach generated over 100K images in total. However, due to issues of visual non-compliance and high similarity among some images, we conducted manual filtering and retained a final set of 90K high-quality images. Additionally, we combine 90K synthetic images and 10K real-life images from DeepFashion~\cite{liu2016deepfashion}.%

\subsection{Multi-view Image Animation and Generation}
\label{sec: mvchamp}

Image-based animating models~\cite{chang2023magicpose, zhu2024champ, hu2024animate, xu2024magicanimate, zhang2024mimicmotion}
struggle to achieve generation of 360$^\circ$-consistent and diverse human videos conditioned on pose sequences (\emph{e.g.}, 2D poses and 3D SMPL(-X) conditions).
To address this limitation, we re-train a state-of-the-art video generative model, Champ~\cite{zhu2024champ}, to obtain a multi-view consistent generative model, MVChamp.
Firstly, We collect a curated dataset of over 100K single-person videos with various motions (\emph{e.g.}, dancing), including approximately 20K videos involving the action of turning around, to fine-tune the model and enhance its generalizability for diverse inputs and motions. 
Secondly, to further improve 3D consistency\cite{han2025vfusion3d}, we utilize scanned models from THuman 2.1 to generate uniformly distributed views via Blender. From these, we select a subset of 24 views that are evenly spaced around the full 360\textdegree\ rotation to fine-tune MVChamp’s temporal layers using diffusion loss.
Lastly, to support high-quality whole-body animation (\emph{i.e.}, body and hands), we introduce precise SMPL-X estimation using NLF~\cite{sarandi2024neural} for body shape and pose estimation. Additionally, we employ HaMeR~\cite{pavlakos2024reconstructing} for hand estimation, rendering the corresponding hand template into depth maps as additional control signals. 

After these training processes, MVChamp generates multi-view images conditioned on SMPL-X parametric model.  The pose condition is set to an “A-Pose” 80\% of the time and a random pose from dance videos (via SMPLer-X \cite{cai2024smpler}) 20\% of the time, balancing pose variability.
To address discrepancies between the first and last frames, we propose a \emph{Temporal Shift Denoising Strategy}: during the denoising steps, we shift latent inputs and pose signals temporally, moving the last frame to the first. This improves continuity between frames without additional inference costs. Finally, FaceFusion~\cite{facefusion} is utilized to enhance facial details.

\begin{figure}
  \centering
  \includegraphics[width=0.9\linewidth]{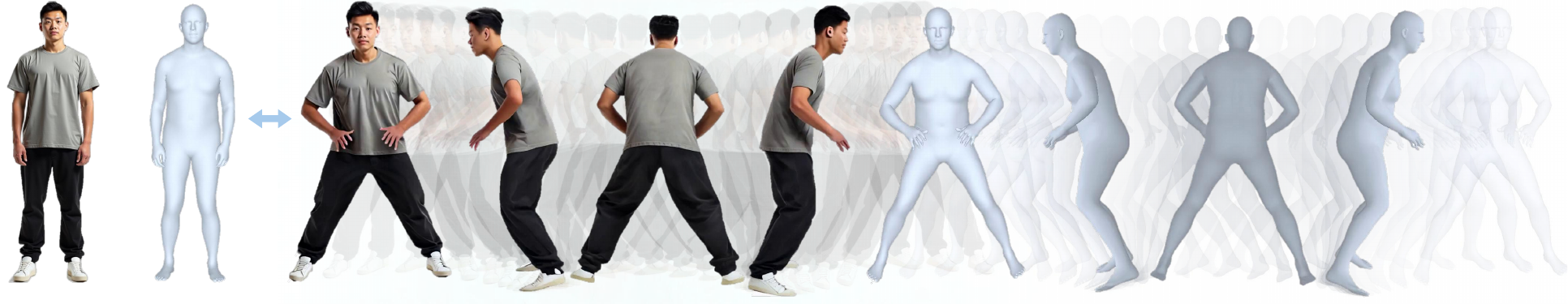}
  \caption{\textbf{A paired example from the proposed \datasetname Dataset.} For each reference image, we generate a set of multi-view images using an estimated shape and a specific pose. The figure shows the pose is well-aligned.
}
  \label{fig:dataset_example}
\end{figure}

\subsection{Data Statistics and Characteristics}
\label{sec: statistic}

Tab.~\ref{tab:datasets} compares our dataset \datasetname to other 3D human datasets regarding scale, diversity, and consistency. With over 100K human identities and 20K poses, \datasetname offers balanced diversity across dimensions such as \textit{area}, \textit{clothing}, \textit{body shape}, \textit{age}, and \textit{gender} (see Fig. \ref{fig:title}). Each sample includes a reference image, SMPL-X estimates, 24 uniformly sampled multi-view images, camera parameters, and SMPL-X data (see Fig. \ref{fig:dataset_example}).

\begin{figure*}[ht]
  \centering
  \includegraphics[width=0.95\linewidth]{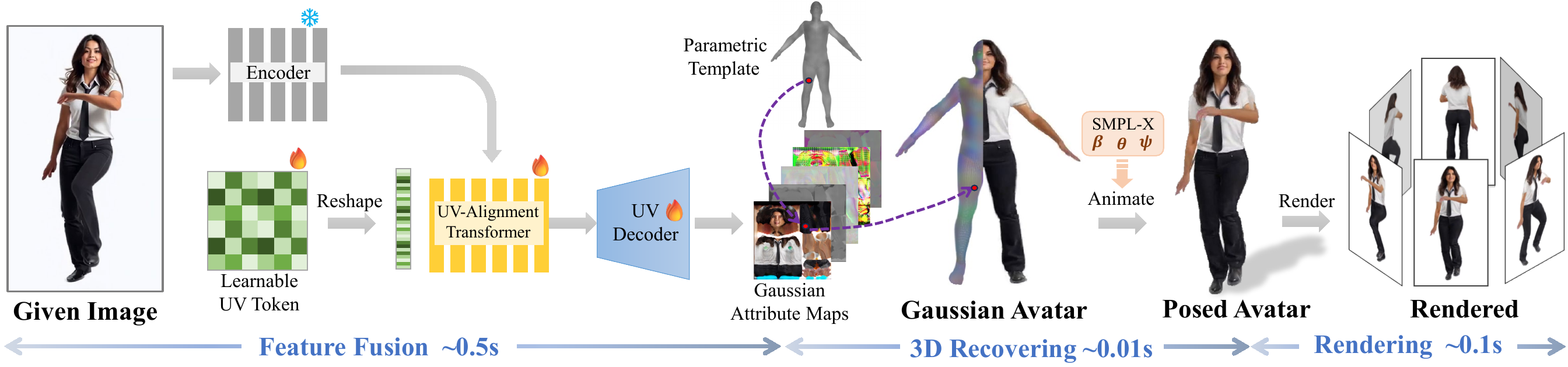}
  \caption{The architecture of \modelname, a full-differentiable transformer-based framework for reconstructing animatable 3D human from a single image. 
  The model integrates a high-resolution ($1024\times1024$) encoder~\cite{khirodkar2025sapiens} and fuses image tokens with learnable UV tokens through the UV-Alignment Transformer.
A UV Decoder predicts Gaussian attribute maps as intermediate representations, capturing the human's geometry and appearance in a structured 2D UV space defined by the SMPL-X model.
 These maps, in conjunction with the SMPL-X model, represent a 3D human avatar in a canonical space, which can be animated using linear blend skinning (LBS). The model is optimized using multi-view images with diverse poses and identities, learning to disentangle pose, appearance, and shape.
 }
  \label{fig:framework}
\end{figure*}

\section{Large Human Reconstruction Model}
\label{sec:method_model}
In this section, we present the large-scale human reconstruction model, named \modelname. The overview pipeline is shown in Fig. \ref{fig:framework}. In Sec. \ref{sec: rep}, we describe the animatable human representation. Sec. \ref{sec: net} details the network architecture of \modelname, while Sec. \ref{sec: training} explains how we trained the model in an end-to-end manner using the multi-view image dataset.

\subsection{Animatable Human Representation}
\label{sec: rep}

Similarly to previous work~\cite{zhang20243gen, hu2024gaussianavatar, kirschstein2024gghead}, \modelname leverages 3D Gaussian Splatting~\cite{kerbl20233d} in conjunction with SMPL-X for 3D human representation, aiming to address the challenges of real-time rendering and accurate animation of human avatars. 
Specifically, each Gaussian primitive $\mathcal{G}_k$ is characterized by:
$
    \mathcal{G}_k = \left\{ \mu_k, \alpha_k, \mathbf{r}_k, \mathbf{s}_k, \mathbf{c}_k \right\},$
where $\mu_k$ is the 3D position of the Gaussian, $\alpha_k$  is opacity, $\mathbf{r}_k$ is the rotation, $\mathbf{s}_k$ is the scale, and $\mathbf{c}_k$ is the color.

\paragraph{3D Gaussian Human.}
Directly predicting all 3D Gaussian primitives is computationally intensive. Instead, \modelname leverages the predefined 2D UV space of the SMPL-X model to transform the 3D representation task into a more manageable 2D problem. Initially, \modelname predicts Gaussian attribute maps that encode the offset values $\{$${\delta_{\mathbf{\mu}_k}, \delta_{\mathbf{r}k}, \delta{\mathbf{s}_k}}$$\}$, as well as the color $\mathbf{c}_k$ and opacity $\alpha_k$ for each Gaussian primitive $\mathcal{G}_k$. 
Similarly to $E^3Gen$~\cite{zhang20243gen}, the position $\mathbf{\mu}_k$, scale $\mathbf{s}_k$, and rotation $\mathbf{r}_k $ of each Gaussian primitive  $\mathcal{G}_k$ is modeled relative to its SMPL-X vertex as follows:
$\mathbf{\mu}_k = \hat{\mathbf{\mu}}_k + \delta_{\mathbf{\mu}_k},
\mathbf{s}_k = \hat{\mathbf{s}}_k \cdot \delta_{\mathbf{s}_k},
\mathbf{r}_k = \hat{\mathbf{r}}_k \cdot \delta_{\mathbf{r}_k}, $
where $\hat{\mathbf{\mu}}_k$, $\hat{\mathbf{s}}_k$, and $\hat{\mathbf{r}}_k$ are the initial values based on the SMPL-X model. The color $\mathbf{c}_k$ is assigned using RGB values, and the opacity $\alpha_k$ is set to $1$, indicating full opacity.
This unified UV space significantly reduces computational complexity and fully exploits the geometric and semantic priors provided by the SMPL-X model. By modeling Gaussian primitives relative to SMPL-X vertices, \modelname ensures semantic consistency across corresponding body parts of diverse avatars, thereby enhancing the model's generalization capability across various reference images.

\paragraph{Animation and Rendering.}
\label{sec:animation}
Given a target pose, we calculate the transformation of each human joint using predefined kinematic relationships. The transformation of each Gaussian primitive is performed using a forward skinning technique based on LBS. Specifically, the position of each Gaussian primitive $\mathcal{G}_k$ is transformed as follows:
$\mathbf{\mu}_k^{\prime} = \sum_{i=1}^{n_b} w_i \mathbf{B}_i \mathbf{\mu}_k$.
Additionally, the rotation matrix $\mathbf{R}_k$ is updated by:
$\mathbf{R}_k^{\prime} = \mathbf{T}_k^{1:3,1:3} \mathbf{R}_k,$
where
$\mathbf{T}_k = \sum_{i=1}^{n_b} w_i \mathbf{B}_i,
$
and $\mathbf{R}_k$ is the rotation matrix representation of the rotation angle $\mathbf{r}_k$. Here, $n_b$ is the number of joints, $\mathbf{B}_i$ is the transformation matrix for each joint, and $w_i$ represents the skinning weights, indicating the influence of each joint's motion on the Gaussian primitive's position $\mathbf{\mu}_k$.


To estimate the skinning weights for each $\mathcal{G}_k$, we first compute a body-part skinning field using a pre-defined low-resolution volume~\cite{chen2023fast}. For smaller regions like the hands and face, which undergo minimal topological variation, weights are interpolated from the SMPL-X template via barycentric coordinates. This strategy enables modeling large topology changes across identities (e.g., from clothing) while stabilizing convergence in less variable areas such as fingers and facial regions.

\subsection{Network Structure}
\label{sec: net}
As illustrated in Fig.~\ref{fig:framework}, \modelname is a full-differentiable framework for reconstructing animatable 3D human. 

\paragraph{High-resolution Image Encoder.}
\label{sec:encoder}
Higher resolution of the input image results in a high-quality reconstruction. However, previous works suffer from low-resolution ViT-based encoders, such as DINOv2~\cite{oquab2023dinov2}, which support the $448\times448$ resolution. To fully leverage the resolution of the \datasetname dataset, we further adopt a high-resolution human foundation model, Sapiens \cite{khirodkar2025sapiens}, to encode the $1024\times1024$ resolution images into patch-wise feature tokens, formulated as:
$
     \mathbf{F} = \{ \mathbf{f}_{i} \}^n_{i=1} \in \mathbb{R}^{d_E},
$
\noindent where $ i $ denotes the $ i $-th image patch, and \( d_E \) is the channel length of each token. 
The Sapiens model is pretrained on 300 million in-the-wild human images using the Masked Autoencoder (MAE) framework~\cite{he2022masked}, enabling it to excel in preserving fine-grained details and capturing diverse human poses and appearances, making it highly effective for high-resolution human image feature extraction.

\paragraph{UV-Alignment Transformer.}

To map irregular and diverse reference images onto regular UV feature maps, a UV-Alignment Transformer is employed to align learnable spatial-positional UV tokens \( \mathbf{Q}^0 \) with reference image features \( \mathbf{F} \). Specifically, the UV-Alignment Transformer concatenates \( \mathbf{F} \) from Spaiens with \( \mathbf{Q}^0 \), and aggregates and refines features through \( D \) transformer blocks, producing an enhanced representation \( \mathbf{Q}^D \). Each transformer block utilizes self-attention mechanisms, enabling the model to capture complex relationships among the input tokens and impute missing parts using correlated tokens.

\paragraph{UV Decoder.}
The UV tokens \( \mathbf{Q}^D \) are reshaped and decoded into Gaussian attribute maps, capturing the human’s geometry and appearance within the structured 2D UV space as shown in Sec. \ref{sec: rep}. 
To preserve and enhance fine details in the decoded Gaussian attribute maps, a Convolutional Neural Network is employed for spatial up-sampling.

\subsection{Training Objectives}
\label{sec: training}

\modelname reconstructs 3D human by predicting Gaussian attribute UV maps in a single forward pass, offering significant advantages in inference speed and enabling end-to-end training with multi-view images.
For each sample, we select a front view as the reference image $\mathbf{I}_{\text{ref}}$, along with a random set of generated multi-view images $\{{\mathbf{I}_{\text{gt},i}}\}_{i=1}^N $ and their corresponding SMPL-X parameters and camera for supervision. It takes $\mathbf{I}_{\text{ref}}$ as input to generate the 3D human, using differentiable rendering to produce the multi-view images $\{{\mathbf{I}_{\text{pred},i}}\}_{i=1}^N $. The loss function is defined as follows:
\begin{equation}
\mathcal{L} = \sum_{i=1}^N \left( \left\lVert \mathbf{I}_{\text{gt},i} - \mathbf{I}_{\text{pred},i} \right\rVert^2  + \lambda L_{vgg}(\mathbf{I}_{\text{gt},i}, \mathbf{I}_{\text{pred}, i})\right),
\end{equation} 
where $\lambda$ controls the balance between the mean square error loss and the perceptual loss \( L_{vgg} \).

\section{Experiments}

\begin{figure*}
  \centering
  \includegraphics[width=\linewidth]{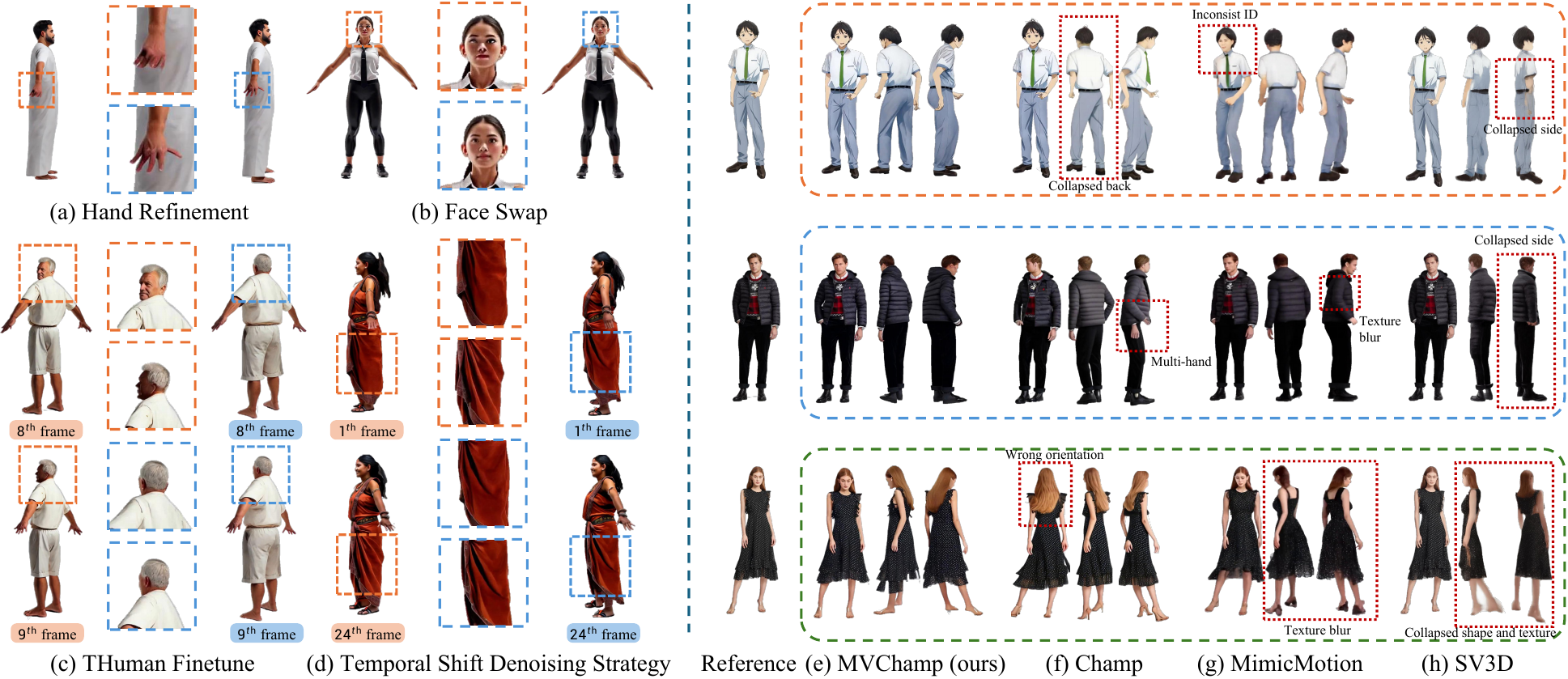}
  \caption{Qualitative results of our MVChamp ablation study (left) and comparison experiment (right).
  }
  \label{fig:comparison_video}
\end{figure*}

\subsection{Experiments on Dataset Creation}
\label{sec: exp_data}

Fig.~\ref{fig:comparison_video} (left) presents the qualitative results of the ablation study on our data generation method. Specifically, Fig.~\ref{fig:comparison_video} (a) and Fig.~\ref{fig:comparison_video} (b) demonstrate how hand refinement and face swapping effectively enhance the quality of the generated hands and faces. Fig.~\ref{fig:comparison_video} (c) highlights the importance of fine-tuning the 3D dataset Thuman 2.1 for improving the 3D consistency of MVChamp, while Fig.~\ref{fig:comparison_video} (d) illustrates that the Temporal Shift Denoising strategy, which involves cycling the first and last frames, improves the consistency between the first and last frames.

Fig.~\ref{fig:comparison_video} (right) compares our MVChamp’s generation quality against previous human image animation models (\emph{e.g.}, Champ, MimicMotion \cite{zhang2024mimicmotion}) and general multi-view video generation model (\emph{e.g.}, SV3D \cite{voleti2025sv3d}). The issues of these methods are highlighted in the figure. Champ struggles to generate plausible hands and heads in multi-view scenarios. MimicMotion not only fails to produce realistic shoes but also has difficulty preserving the identity, especially for the anime image. SV3D produces low-quality multi-view human generation due to the lack of 3D prior knowledge of the human body. In contrast, our model generates consistent, high-quality results across views.

\subsection{Comparison of Reconstruction Model}
\label{sec: exp_comp}

\paragraph{Implementation Details.}
For the encoder, we employ the pre-trained Sapiens-1B model~\cite{khirodkar2025sapiens} and keep its weights frozen during training. We then define a transformer-based framework with a parameter size of $0.5B$ to perform feature fusion. Additionally, we densify the SMPL-X vertex set by sampling approximately $200,000$ vertices to map attributes and represent the 3D human model effectively.

\paragraph{Dataset and Metrics.} 
We train \modelname on a dataset consisting of generated multi-view images from \datasetname and rendered images from THuman 2.1~\cite{zheng2019deephuman}. 
For THuman 2.1, we render 72 view images at a resolution of $896\times640$ for each scan, with rendering views uniformly distributed.
To evaluate performance quantitatively, we reserve the last 50 cases from both \datasetname and THuman 2.1 as the testing set. 
We use the most frontal image as the reference image and the remaining images as ground-truth data for evaluation.  The ground-truth camera parameters and SMPL-X parameters are provided for all methods. 
The renderings of the reconstructed model are then compared to their corresponding ground truth images with several metrics, including Mean Squared Error (MSE), Learned Perceptual Image Patch Similarity (LPIPS) \cite{zhang2018unreasonable}, Peak Signal-to-Noise Ratio (PSNR). \emph{More details on implementation can be found in the supplementary material.}

\begin{figure*}
  \centering
  \includegraphics[width= 1\linewidth]{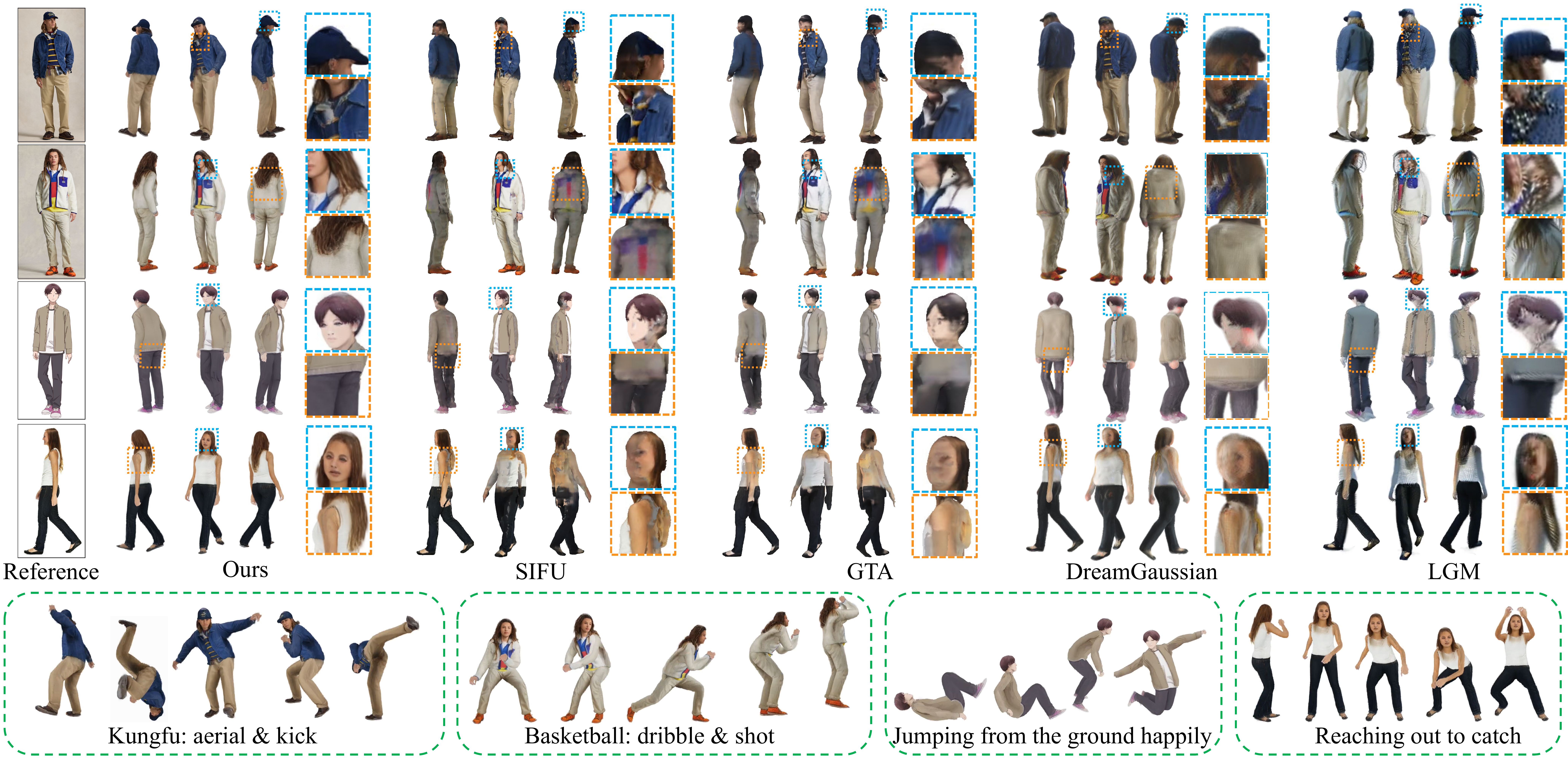}
  \caption{Comparisons on (a) the upper: novel-view synthesis given a single image, and (b) the lower: our animated results.}
  \label{fig:comparison_3D}
\end{figure*}

\paragraph{Baselines.} 
We compare IDOL with three baseline categories in the Single-Image Human Reconstruction task. The first category comprises methods based on loop optimization or pixel-alignment modules, including GTA~\cite{zhang2024global} and SIFU~\cite{zhang2024sifu}. These methods focus on refining human reconstruction through iterative optimization or predicting 3D geometry from pixel-aligned features.

The second category encompasses large-scale generic reconstruction networks, represented by LGM~\cite{tang2025lgm}. These models are known for their ability to handle large datasets and their scalability, offering advantages in terms of fast reference speeds and large output resolutions. %

The third category involves optimization-based 3D generation methods using score distillation sampling (SDS), exemplified by DreamGaussian~\cite{tang2023dreamgaussian}. These methods leverage priors from 2D diffusion models to distill 3D objects. DreamGaussian accelerates convergence by progressively densifying 3D Gaussians, significantly reducing the reconstruction time. However, it still takes approximately two minutes to reconstruct a single object.

\paragraph{Quantitative comparison.} 
As shown in Tab~\ref{tab:comparison_3d}, our method outperforms all baselines in all metrics. We attribute this superior performance to the large-scale dataset HuGe100K and the design of our large-scale reconstruction model IDOL, which allows for more effective training and improved synthesis of appearance results. 
Despite this, we note that SIFU and GTA report a lower metric than what we expected in our test settings. While we provide accurate ground-truth SMPL-X parameters and camera settings, the ideal orthographic projection required by the pixel-alignment modules in SIFU and GTA is not well-suited to our perspective projection model, where the camera focus ranges from $35$ to $80$. This mismatch leads to a misalignment between the rendered images and the ground truth, adversely affecting their performance metrics.  In fact, many real-life photographs are taken with medium to short focal lengths and cannot be approximated using orthographic projection, which is a less noticeable drawback of SIFU and GTA.
Additionally, SIFU and GTA, trained with images at a resolution of $512 \times 512$, struggle to synthesize detailed textures, particularly in the invisible areas of the reference images. This limitation is primarily due to the lack of comprehensiveness and diversity in their training datasets, which restricts their performance in generating invisible aspects of the images.

\paragraph{Qualitative comparison.}
We perform a qualitative evaluation on an in-the-wild dataset with methods from \cite{zhang2024sifu, zhang2024global, tang2023dreamgaussian, tang2025lgm}, with the results presented in Fig. \ref{fig:comparison_3D}.
Our evaluation includes a subject in a complex outfit featuring a baseball cap and textured clothing, demonstrating our method’s ability to synthesize intricate textures and handle loose outfits across different views. Additional tests include out-of-domain cartoon data and large-angle side views and assessing model adaptability and viewpoint handling. IDOL consistently outperforms the baselines, which struggle with detail reproduction and texture consistency under these various conditions.

\subsection{Ablation Study on Reconstruction Model} 
\label{sec: exp_abl}

\begin{figure}
  \centering
  \includegraphics[width=0.9\linewidth]{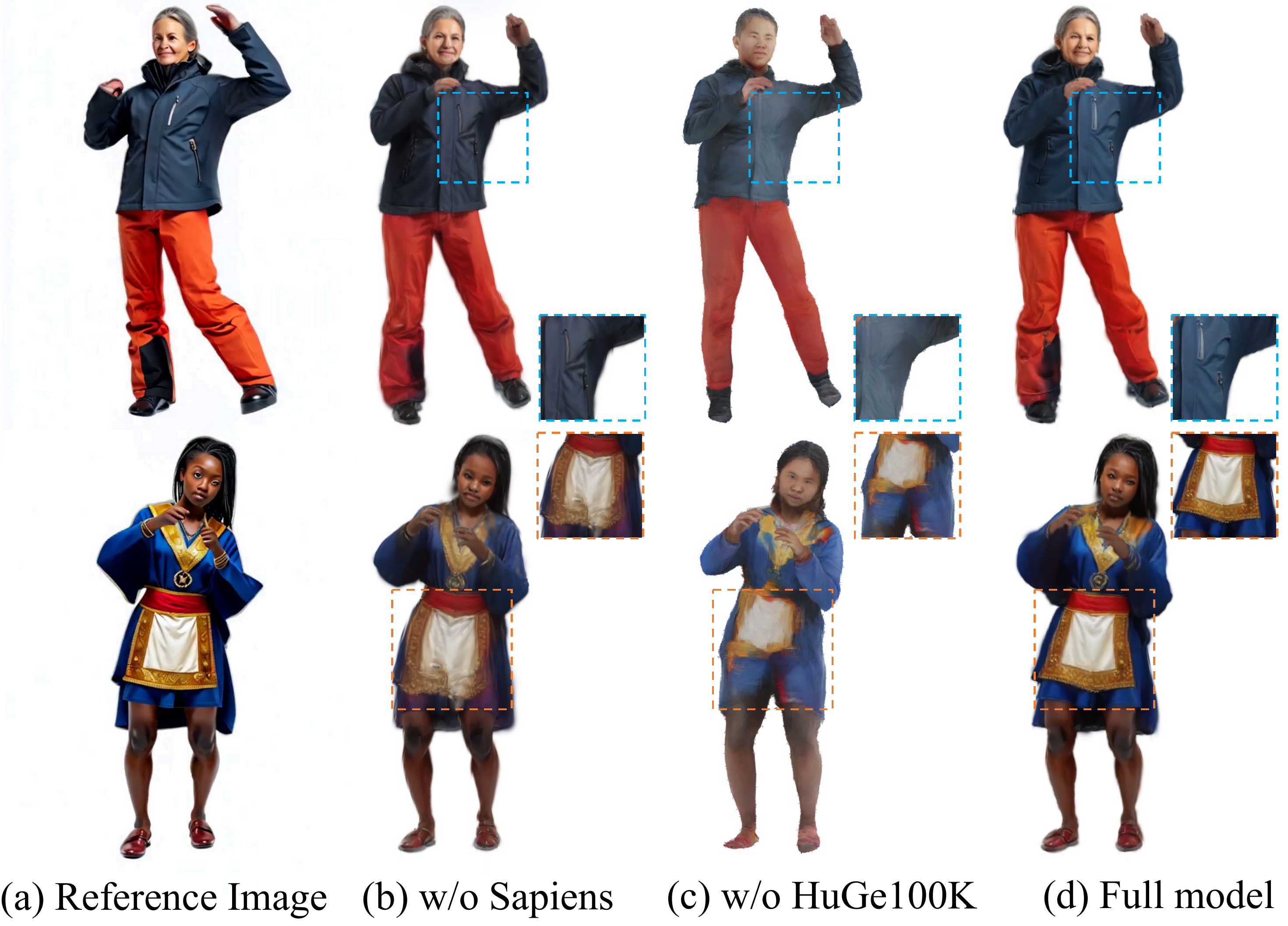}
  \caption{Qualitative Results of Ablation Study of IDOL.}
  \label{tab:abl_study}
\end{figure}

We assess the impact of proposed components by removing them individually from IDOL. As shown in Fig. \ref{tab:abl_study}, replacing the Sapiens~\cite{khirodkar2025sapiens} encoder with DINO v2~\cite{oquab2023dinov2} (w/o Sapiens) reduces detail quality, resulting in less realistic textures and folds. Excluding the \datasetname dataset (w/o \datasetname) causes significant distortion, including blurred details and color bleeding. The complete model, with all components, produces the most realistic and detailed results. These findings highlight the critical roles of both the Sapiens encoder and the \datasetname dataset in achieving high-quality avatar generation.

\begin{table}[t]
\centering
\resizebox{0.4\textwidth}{!}{
    \begin{tabular}{lccc}
    \toprule
    Method & MSE $\downarrow$ & PSNR $\uparrow$ & LPIPS $\downarrow$ \\
    \midrule
    SIFU~\cite{zhang2024sifu} & 0.042 & 14.204  & 1.612 \\ 
    GTA~\cite{zhang2024global}  & 0.041 & 14.282  & 1.629 \\ 
    Ours-w/o \datasetname & 0.017 & 19.225  & 1.326 \\
    Ours-full & \textbf{0.008} & \textbf{21.673} & \textbf{1.138} \\ 
    \bottomrule
    \end{tabular}}
\caption{Evaluation of Comparison and Ablation Experiments.}
\label{tab:comparison_3d}
\end{table}

\subsection{Downstream Applications}
\label{sec: exp_ds}

\begin{figure}
  \centering
  \includegraphics[width=\linewidth]{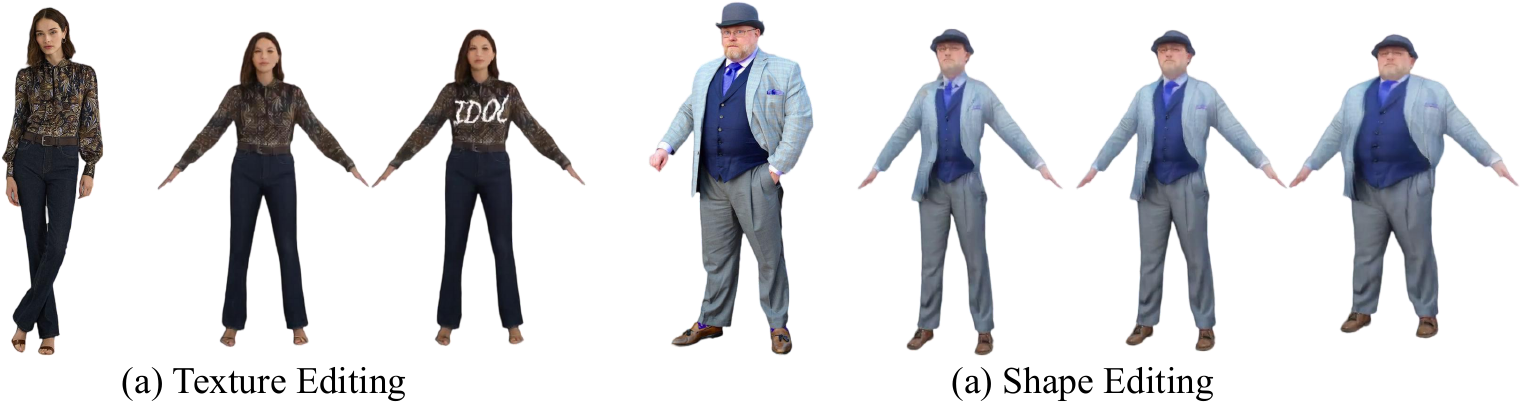}
  \caption{Controllable Avatar Editing: (a) texture editing; (b) body shape editing.}
  \label{fig:applications}
\end{figure}

\modelname reconstructs an avatar by combining Gaussian attribute maps with the SMPL-X model, enabling users to modify the avatar's appearance by editing UV texture maps and to control body shape by adjusting SMPL-X shape parameters. Fig. \ref{fig:applications}(a) illustrates the effect of texture editing on clothing patterns, while Fig. \ref{fig:applications}(b) demonstrates shape editing to adjust the avatar's body size. This approach provides a high degree of controllability over both the avatar's appearance and body shape.

\section{Conclusions and Limitations}
In conclusion, our work has made significant strides in creating an animatable 3D human from a single image. We introduced a scalable pipeline for training a simple yet efficient feed-forward model, incorporating a dataset generation framework, a large-scale dataset \datasetname, and a scalable reconstruction transformer model, \modelname. This model efficiently reconstructs photorealistic 3D humans in under a second and demonstrates versatility across various applications.
This model can efficiently reconstruct photorealistic humans in less than a second and is versatile enough to support various applications.

However, there are limitations. Due to the constraints of the video model used, we can only synthesize single-frame images from fixed viewpoints. Future work could consider generating longer motion sequences. 
 The focus on body reconstruction and animation leaves facial optimization secondary, and the architecture lacks specific design for facial identity or expression. 
 Additionally, handling half-body inputs remains challenging, and improvements in data generation strategies are needed to enhance performance.

\section*{Acknowledgments}
This work was partly supported by the National Key Research and Development Program of China (2022YFF0902400), the Shenzhen Science and Technology Program (JCYJ20220818101014030), and the research fund of Tsinghua University - Tencent Joint Laboratory for Internet Innovation Technology.

{
    \small
    \bibliographystyle{ieeenat_fullname}
    \bibliography{main}

\begin{thebibliography}{93}
\providecommand{\natexlab}[1]{#1}
\providecommand{\url}[1]{\texttt{#1}}
\expandafter\ifx\csname urlstyle\endcsname\relax
  \providecommand{\doi}[1]{doi: #1}\else
  \providecommand{\doi}{doi: \begingroup \urlstyle{rm}\Url}\fi

\bibitem[Achiam et~al.(2023)Achiam, Adler, Agarwal, Ahmad, Akkaya, Aleman, Almeida, Altenschmidt, Altman, Anadkat, et~al.]{achiam2023gpt}
Josh Achiam, Steven Adler, Sandhini Agarwal, Lama Ahmad, Ilge Akkaya, Florencia~Leoni Aleman, Diogo Almeida, Janko Altenschmidt, Sam Altman, Shyamal Anadkat, et~al.
\newblock Gpt-4 technical report.
\newblock \emph{arXiv preprint arXiv:2303.08774}, 2023.

\bibitem[AlBahar et~al.(2023)AlBahar, Saito, Tseng, Kim, Kopf, and Huang]{HumanSGD:2023}
Badour AlBahar, Shunsuke Saito, Hung-Yu Tseng, Changil Kim, Johannes Kopf, and Jia-Bin Huang.
\newblock Single-image 3d human digitization with shape-guided diffusion.
\newblock In \emph{SIGGRAPH Asia 2023 Conference Papers}, 2023.

\bibitem[Blattmann et~al.(2023)Blattmann, Dockhorn, Kulal, Mendelevitch, Kilian, Lorenz, Levi, English, Voleti, Letts, et~al.]{blattmann2023svd}
Andreas Blattmann, Tim Dockhorn, Sumith Kulal, Daniel Mendelevitch, Maciej Kilian, Dominik Lorenz, Yam Levi, Zion English, Vikram Voleti, Adam Letts, et~al.
\newblock Stable video diffusion: Scaling latent video diffusion models to large datasets.
\newblock \emph{arXiv preprint arXiv:2311.15127}, 2023.

\bibitem[Cai et~al.(2022)Cai, Ren, Zeng, Lin, Yu, Wang, Fan, Gao, Yu, Pan, et~al.]{cai2022humman}
Zhongang Cai, Daxuan Ren, Ailing Zeng, Zhengyu Lin, Tao Yu, Wenjia Wang, Xiangyu Fan, Yang Gao, Yifan Yu, Liang Pan, et~al.
\newblock Humman: Multi-modal 4d human dataset for versatile sensing and modeling.
\newblock In \emph{European Conference on Computer Vision}, pages 557--577. Springer, 2022.

\bibitem[Cai et~al.(2024)Cai, Yin, Zeng, Wei, Sun, Yanjun, Pang, Mei, Zhang, Zhang, et~al.]{cai2024smpler}
Zhongang Cai, Wanqi Yin, Ailing Zeng, Chen Wei, Qingping Sun, Wang Yanjun, Hui~En Pang, Haiyi Mei, Mingyuan Zhang, Lei Zhang, et~al.
\newblock Smpler-x: Scaling up expressive human pose and shape estimation.
\newblock \emph{Advances in Neural Information Processing Systems}, 36, 2024.

\bibitem[Chang et~al.(2023)Chang, Shi, Gao, Xu, Fu, Song, Yan, Zhu, Yang, and Soleymani]{chang2023magicpose}
Di Chang, Yichun Shi, Quankai Gao, Hongyi Xu, Jessica Fu, Guoxian Song, Qing Yan, Yizhe Zhu, Xiao Yang, and Mohammad Soleymani.
\newblock Magicpose: Realistic human poses and facial expressions retargeting with identity-aware diffusion.
\newblock In \emph{Forty-first International Conference on Machine Learning}, 2023.

\bibitem[Chen et~al.(2023)Chen, Jiang, Song, Rietmann, Geiger, Black, and Hilliges]{chen2023fast}
Xu Chen, Tianjian Jiang, Jie Song, Max Rietmann, Andreas Geiger, Michael~J Black, and Otmar Hilliges.
\newblock Fast-snarf: A fast deformer for articulated neural fields.
\newblock \emph{IEEE Transactions on Pattern Analysis and Machine Intelligence}, 45\penalty0 (10):\penalty0 11796--11809, 2023.

\bibitem[Cheng et~al.(2022)Cheng, Xu, Piao, Qian, Wu, Lin, and Li]{cheng2022generalizable}
Wei Cheng, Su Xu, Jingtan Piao, Chen Qian, Wayne Wu, Kwan-Yee Lin, and Hongsheng Li.
\newblock Generalizable neural performer: Learning robust radiance fields for human novel view synthesis.
\newblock \emph{arXiv preprint arXiv:2204.11798}, 2022.

\bibitem[Cheng et~al.(2023)Cheng, Chen, Fan, Yin, Chen, Cai, Wang, Gao, Yu, Lin, et~al.]{cheng2023dna}
Wei Cheng, Ruixiang Chen, Siming Fan, Wanqi Yin, Keyu Chen, Zhongang Cai, Jingbo Wang, Yang Gao, Zhengming Yu, Zhengyu Lin, et~al.
\newblock Dna-rendering: A diverse neural actor repository for high-fidelity human-centric rendering.
\newblock In \emph{Proceedings of the IEEE/CVF International Conference on Computer Vision}, pages 19982--19993, 2023.

\bibitem[Corona et~al.(2023)Corona, Zanfir, Alldieck, Bazavan, Zanfir, and Sminchisescu]{corona2023structured}
Enric Corona, Mihai Zanfir, Thiemo Alldieck, Eduard~Gabriel Bazavan, Andrei Zanfir, and Cristian Sminchisescu.
\newblock Structured 3d features for reconstructing controllable avatars.
\newblock In \emph{Proceedings of the IEEE/CVF Conference on Computer Vision and Pattern Recognition}, pages 16954--16964, 2023.

\bibitem[Deitke et~al.(2023)Deitke, Schwenk, Salvador, Weihs, Michel, VanderBilt, Schmidt, Ehsani, Kembhavi, and Farhadi]{deitke2023objaverse}
Matt Deitke, Dustin Schwenk, Jordi Salvador, Luca Weihs, Oscar Michel, Eli VanderBilt, Ludwig Schmidt, Kiana Ehsani, Aniruddha Kembhavi, and Ali Farhadi.
\newblock Objaverse: A universe of annotated 3d objects.
\newblock In \emph{Proceedings of the IEEE/CVF Conference on Computer Vision and Pattern Recognition}, pages 13142--13153, 2023.

\bibitem[Deitke et~al.(2024)Deitke, Liu, Wallingford, Ngo, Michel, Kusupati, Fan, Laforte, Voleti, Gadre, et~al.]{deitke2024objaverse-xl}
Matt Deitke, Ruoshi Liu, Matthew Wallingford, Huong Ngo, Oscar Michel, Aditya Kusupati, Alan Fan, Christian Laforte, Vikram Voleti, Samir~Yitzhak Gadre, et~al.
\newblock Objaverse-xl: A universe of 10m+ 3d objects.
\newblock \emph{Advances in Neural Information Processing Systems}, 36, 2024.

\bibitem[Feng et~al.(2022)Feng, Liu, Lai, Yang, and Li]{li2022neurips}
Qiao Feng, Yebin Liu, Yu-Kun Lai, Jingyu Yang, and Kun Li.
\newblock Fof: Learning fourier occupancy field for monocular real-time human reconstruction.
\newblock In \emph{NeurIPS}, 2022.

\bibitem[Gao et~al.(2022)Gao, Yang, Kim, Peng, Liu, and Tong]{gao2022mps}
Xiangjun Gao, Jiaolong Yang, Jongyoo Kim, Sida Peng, Zicheng Liu, and Xin Tong.
\newblock Mps-nerf: Generalizable 3d human rendering from multiview images.
\newblock \emph{IEEE Transactions on Pattern Analysis and Machine Intelligence}, 2022.

\bibitem[Gao et~al.(2024{\natexlab{a}})Gao, Li, Zhang, Zhang, Cao, Shan, and Quan]{gao2024contex}
Xiangjun Gao, Xiaoyu Li, Chaopeng Zhang, Qi Zhang, Yanpei Cao, Ying Shan, and Long Quan.
\newblock Contex-human: Free-view rendering of human from a single image with texture-consistent synthesis.
\newblock In \emph{Proceedings of the IEEE/CVF Conference on Computer Vision and Pattern Recognition}, pages 10084--10094, 2024{\natexlab{a}}.

\bibitem[Gao et~al.(2024{\natexlab{b}})Gao, Li, Zhuang, Zhang, Hu, Zhang, Yao, Shan, and Quan]{gao2024mani}
Xiangjun Gao, Xiaoyu Li, Yiyu Zhuang, Qi Zhang, Wenbo Hu, Chaopeng Zhang, Yao Yao, Ying Shan, and Long Quan.
\newblock Mani-gs: Gaussian splatting manipulation with triangular mesh.
\newblock \emph{arXiv preprint arXiv:2405.17811}, 2024{\natexlab{b}}.

\bibitem[Guo et~al.(2024)Guo, Xiang, Ma, Zhou, Li, and Zhang]{guo2024make}
Zhiyang Guo, Jinxu Xiang, Kai Ma, Wengang Zhou, Houqiang Li, and Ran Zhang.
\newblock Make-it-animatable: An efficient framework for authoring animation-ready 3d characters.
\newblock \emph{arXiv preprint arXiv:2411.18197}, 2024.

\bibitem[Han et~al.(2025)Han, Kokkinos, and Torr]{han2025vfusion3d}
Junlin Han, Filippos Kokkinos, and Philip Torr.
\newblock Vfusion3d: Learning scalable 3d generative models from video diffusion models.
\newblock In \emph{European Conference on Computer Vision}, pages 333--350. Springer, 2025.

\bibitem[Han et~al.(2023)Han, Park, Yoon, Kang, Park, and Jeon]{han2023high}
Sang-Hun Han, Min-Gyu Park, Ju~Hong Yoon, Ju-Mi Kang, Young-Jae Park, and Hae-Gon Jeon.
\newblock High-fidelity 3d human digitization from single 2k resolution images.
\newblock In \emph{Proceedings of the IEEE/CVF Conference on Computer Vision and Pattern Recognition}, pages 12869--12879, 2023.

\bibitem[He et~al.(2022)He, Chen, Xie, Li, Doll{\'a}r, and Girshick]{he2022masked}
Kaiming He, Xinlei Chen, Saining Xie, Yanghao Li, Piotr Doll{\'a}r, and Ross Girshick.
\newblock Masked autoencoders are scalable vision learners.
\newblock In \emph{Proceedings of the IEEE/CVF conference on computer vision and pattern recognition}, pages 16000--16009, 2022.

\bibitem[He et~al.(2021)He, Xu, Saito, Soatto, and Tung]{he2021arch++}
Tong He, Yuanlu Xu, Shunsuke Saito, Stefano Soatto, and Tony Tung.
\newblock Arch++: Animation-ready clothed human reconstruction revisited.
\newblock In \emph{Proceedings of the IEEE/CVF international conference on computer vision}, pages 11046--11056, 2021.

\bibitem[He et~al.(2024{\natexlab{a}})He, Li, Kang, Ye, Zhang, Chen, Gao, Zhang, Wu, and Zhuang]{he2024magicman}
Xu He, Xiaoyu Li, Di Kang, Jiangnan Ye, Chaopeng Zhang, Liyang Chen, Xiangjun Gao, Han Zhang, Zhiyong Wu, and Haolin Zhuang.
\newblock Magicman: Generative novel view synthesis of humans with 3d-aware diffusion and iterative refinement.
\newblock \emph{arXiv preprint arXiv:2408.14211}, 2024{\natexlab{a}}.

\bibitem[He et~al.(2024{\natexlab{b}})He, Zhuang, Wang, Yao, Zhu, Li, Zhang, Cao, and Zhu]{he2024head360}
Yuxiao He, Yiyu Zhuang, Yanwen Wang, Yao Yao, Siyu Zhu, Xiaoyu Li, Qi Zhang, Xun Cao, and Hao Zhu.
\newblock Head360: Learning a parametric 3d full-head for free-view synthesis in 360.
\newblock In \emph{European Conference on Computer Vision}, pages 254--272. Springer, 2024{\natexlab{b}}.

\bibitem[Hong et~al.(2023)Hong, Zhang, Gu, Bi, Zhou, Liu, Liu, Sunkavalli, Bui, and Tan]{hong2023lrm}
Yicong Hong, Kai Zhang, Jiuxiang Gu, Sai Bi, Yang Zhou, Difan Liu, Feng Liu, Kalyan Sunkavalli, Trung Bui, and Hao Tan.
\newblock Lrm: Large reconstruction model for single image to 3d.
\newblock \emph{arXiv preprint arXiv:2311.04400}, 2023.

\bibitem[https://github.com/black-forest labs/flux(2024)]{flux}
https://github.com/black-forest labs/flux.
\newblock Flux latent rectified flow transformers, 2024.

\bibitem[https://github.com/facefusion/facefusion(2024)]{facefusion}
https://github.com/facefusion/facefusion.
\newblock Facefusion, 2024.

\bibitem[https://renderpeople.com/3d people(2015)]{renderpeople}
https://renderpeople.com/3d people.
\newblock Renderpeople dataset, 2015.

\bibitem[https://web.twindom.com/(2020)]{twindom}
https://web.twindom.com/.
\newblock Twindom dataset, 2020.

\bibitem[Hu(2024)]{hu2024animate}
Li Hu.
\newblock Animate anyone: Consistent and controllable image-to-video synthesis for character animation.
\newblock In \emph{Proceedings of the IEEE/CVF Conference on Computer Vision and Pattern Recognition}, pages 8153--8163, 2024.

\bibitem[Hu et~al.(2024)Hu, Zhang, Zhang, Zhou, Liu, Zhang, and Nie]{hu2024gaussianavatar}
Liangxiao Hu, Hongwen Zhang, Yuxiang Zhang, Boyao Zhou, Boning Liu, Shengping Zhang, and Liqiang Nie.
\newblock Gaussianavatar: Towards realistic human avatar modeling from a single video via animatable 3d gaussians.
\newblock In \emph{Proceedings of the IEEE/CVF Conference on Computer Vision and Pattern Recognition}, pages 634--644, 2024.

\bibitem[Hu et~al.(2023)Hu, Hong, Pan, Mei, Yang, and Liu]{hu2023sherf}
Shoukang Hu, Fangzhou Hong, Liang Pan, Haiyi Mei, Lei Yang, and Ziwei Liu.
\newblock Sherf: Generalizable human nerf from a single image.
\newblock In \emph{Proceedings of the IEEE/CVF International Conference on Computer Vision (ICCV)}, 2023.

\bibitem[Huang et~al.(2023)Huang, Yi, Liu, Wang, Wu, Wang, Lin, Zhang, and Cai]{huang2022elicit}
Yangyi Huang, Hongwei Yi, Weiyang Liu, Haofan Wang, Boxi Wu, Wenxiao Wang, Binbin Lin, Debing Zhang, and Deng Cai.
\newblock One-shot implicit animatable avatars with model-based priors.
\newblock In \emph{IEEE Conference on Computer Vision (ICCV)}, 2023.

\bibitem[Huang et~al.(2024)Huang, Wang, Zeng, Zha, Zhang, and Liu]{huang2024dreamwaltz-g}
Yukun Huang, Jianan Wang, Ailing Zeng, Zheng-Jun Zha, Lei Zhang, and Xihui Liu.
\newblock {DreamWaltz-G: Expressive 3D Gaussian Avatars from Skeleton-Guided 2D Diffusion}.
\newblock 2024.

\bibitem[Huang et~al.(2020)Huang, Xu, Lassner, Li, and Tung]{huang2020arch}
Zeng Huang, Yuanlu Xu, Christoph Lassner, Hao Li, and Tony Tung.
\newblock Arch: Animatable reconstruction of clothed humans.
\newblock In \emph{Proceedings of the IEEE/CVF Conference on Computer Vision and Pattern Recognition}, pages 3093--3102, 2020.

\bibitem[I\c{s}{\i}k et~al.(2023)I\c{s}{\i}k, Rünz, Georgopoulos, Khakhulin, Starck, Agapito, and Nießner]{isik2023humanrf}
Mustafa I\c{s}{\i}k, Martin Rünz, Markos Georgopoulos, Taras Khakhulin, Jonathan Starck, Lourdes Agapito, and Matthias Nießner.
\newblock Humanrf: High-fidelity neural radiance fields for humans in motion.
\newblock \emph{ACM Transactions on Graphics (TOG)}, 42\penalty0 (4):\penalty0 1--12, 2023.

\bibitem[Kerbl et~al.(2023)Kerbl, Kopanas, Leimk{\"u}hler, and Drettakis]{kerbl20233d}
Bernhard Kerbl, Georgios Kopanas, Thomas Leimk{\"u}hler, and George Drettakis.
\newblock 3d gaussian splatting for real-time radiance field rendering.
\newblock \emph{ACM Trans. Graph.}, 42\penalty0 (4):\penalty0 139--1, 2023.

\bibitem[Khirodkar et~al.(2025)Khirodkar, Bagautdinov, Martinez, Zhaoen, James, Selednik, Anderson, and Saito]{khirodkar2025sapiens}
Rawal Khirodkar, Timur Bagautdinov, Julieta Martinez, Su Zhaoen, Austin James, Peter Selednik, Stuart Anderson, and Shunsuke Saito.
\newblock Sapiens: Foundation for human vision models.
\newblock In \emph{European Conference on Computer Vision}, pages 206--228. Springer, 2025.

\bibitem[Kirillov et~al.(2023)Kirillov, Mintun, Ravi, Mao, Rolland, Gustafson, Xiao, Whitehead, Berg, Lo, et~al.]{kirillov2023segment}
Alexander Kirillov, Eric Mintun, Nikhila Ravi, Hanzi Mao, Chloe Rolland, Laura Gustafson, Tete Xiao, Spencer Whitehead, Alexander~C Berg, Wan-Yen Lo, et~al.
\newblock Segment anything.
\newblock In \emph{Proceedings of the IEEE/CVF International Conference on Computer Vision}, pages 4015--4026, 2023.

\bibitem[Kirschstein et~al.(2024)Kirschstein, Giebenhain, Tang, Georgopoulos, and Nie{\ss}ner]{kirschstein2024gghead}
Tobias Kirschstein, Simon Giebenhain, Jiapeng Tang, Markos Georgopoulos, and Matthias Nie{\ss}ner.
\newblock Gghead: Fast and generalizable 3d gaussian heads.
\newblock \emph{arXiv preprint arXiv:2406.09377}, 2024.

\bibitem[Li et~al.(2024{\natexlab{a}})Li, Tan, Zhang, Xu, Luan, Xu, Hong, Sunkavalli, Shakhnarovich, and Bi]{li2023instant3d}
Jiahao Li, Hao Tan, Kai Zhang, Zexiang Xu, Fujun Luan, Yinghao Xu, Yicong Hong, Kalyan Sunkavalli, Greg Shakhnarovich, and Sai Bi.
\newblock Instant3d: Fast text-to-3d with sparse-view generation and large reconstruction model.
\newblock In \emph{International Conference on Learning Representations (ICLR)}, 2024{\natexlab{a}}.

\bibitem[Li et~al.(2024{\natexlab{b}})Li, Zheng, Liu, Yu, Li, Qi, Li, Chi, Xia, Xue, et~al.]{li2024pshuman}
Peng Li, Wangguandong Zheng, Yuan Liu, Tao Yu, Yangguang Li, Xingqun Qi, Mengfei Li, Xiaowei Chi, Siyu Xia, Wei Xue, et~al.
\newblock Pshuman: Photorealistic single-view human reconstruction using cross-scale diffusion.
\newblock \emph{arXiv preprint arXiv:2409.10141}, 2024{\natexlab{b}}.

\bibitem[Li et~al.(2021)Li, Yang, Ross, and Kanazawa]{li2021ai}
Ruilong Li, Shan Yang, David~A Ross, and Angjoo Kanazawa.
\newblock Ai choreographer: Music conditioned 3d dance generation with aist++.
\newblock In \emph{Proceedings of the IEEE/CVF International Conference on Computer Vision}, pages 13401--13412, 2021.

\bibitem[Li et~al.(2024{\natexlab{c}})Li, Zheng, Wang, and Liu]{li2024animatable}
Zhe Li, Zerong Zheng, Lizhen Wang, and Yebin Liu.
\newblock Animatable gaussians: Learning pose-dependent gaussian maps for high-fidelity human avatar modeling.
\newblock In \emph{Proceedings of the IEEE/CVF Conference on Computer Vision and Pattern Recognition}, pages 19711--19722, 2024{\natexlab{c}}.

\bibitem[Lin et~al.(2023)Lin, Zeng, Lu, Cai, Zhang, Wang, and Zhang]{lin2023motion}
Jing Lin, Ailing Zeng, Shunlin Lu, Yuanhao Cai, Ruimao Zhang, Haoqian Wang, and Lei Zhang.
\newblock Motion-x: A large-scale 3d expressive whole-body human motion dataset.
\newblock \emph{Advances in Neural Information Processing Systems}, 36:\penalty0 25268--25280, 2023.

\bibitem[Liu et~al.(2021)Liu, Habermann, Rudnev, Sarkar, Gu, and Theobalt]{liu2021neural}
Lingjie Liu, Marc Habermann, Viktor Rudnev, Kripasindhu Sarkar, Jiatao Gu, and Christian Theobalt.
\newblock Neural actor: Neural free-view synthesis of human actors with pose control.
\newblock \emph{ACM transactions on graphics (TOG)}, 40\penalty0 (6):\penalty0 1--16, 2021.

\bibitem[Liu et~al.(2024)Liu, Cun, Liu, Wang, Zhang, Chen, Liu, Zeng, Chan, and Shan]{liu2024evalcrafter}
Yaofang Liu, Xiaodong Cun, Xuebo Liu, Xintao Wang, Yong Zhang, Haoxin Chen, Yang Liu, Tieyong Zeng, Raymond Chan, and Ying Shan.
\newblock Evalcrafter: Benchmarking and evaluating large video generation models.
\newblock In \emph{Proceedings of the IEEE/CVF Conference on Computer Vision and Pattern Recognition}, pages 22139--22149, 2024.

\bibitem[Liu et~al.(2016)Liu, Luo, Qiu, Wang, and Tang]{liu2016deepfashion}
Ziwei Liu, Ping Luo, Shi Qiu, Xiaogang Wang, and Xiaoou Tang.
\newblock Deepfashion: Powering robust clothes recognition and retrieval with rich annotations.
\newblock In \emph{Proceedings of the IEEE conference on computer vision and pattern recognition}, pages 1096--1104, 2016.

\bibitem[Loper et~al.(2015)Loper, Mahmood, Romero, Pons-Moll, and Black]{loper2015smpl}
Matthew Loper, Naureen Mahmood, Javier Romero, Gerard Pons-Moll, and Michael~J Black.
\newblock Smpl: A skinned multi-person linear model.
\newblock \emph{ACM Transactions on Graphics}, 34\penalty0 (6), 2015.

\bibitem[Luo et~al.(2024)Luo, Rockwell, Lee, and Johnson]{luo2024cap3d}
Tiange Luo, Chris Rockwell, Honglak Lee, and Justin Johnson.
\newblock Scalable 3d captioning with pretrained models.
\newblock \emph{Advances in Neural Information Processing Systems}, 36, 2024.

\bibitem[Mildenhall et~al.(2021)Mildenhall, Srinivasan, Tancik, Barron, Ramamoorthi, and Ng]{mildenhall2021nerf}
Ben Mildenhall, Pratul~P Srinivasan, Matthew Tancik, Jonathan~T Barron, Ravi Ramamoorthi, and Ren Ng.
\newblock Nerf: Representing scenes as neural radiance fields for view synthesis.
\newblock \emph{Communications of the ACM}, 65\penalty0 (1):\penalty0 99--106, 2021.

\bibitem[Oquab et~al.(2023)Oquab, Darcet, Moutakanni, Vo, Szafraniec, Khalidov, Fernandez, Haziza, Massa, El-Nouby, et~al.]{oquab2023dinov2}
Maxime Oquab, Timoth{\'e}e Darcet, Th{\'e}o Moutakanni, Huy Vo, Marc Szafraniec, Vasil Khalidov, Pierre Fernandez, Daniel Haziza, Francisco Massa, Alaaeldin El-Nouby, et~al.
\newblock Dinov2: Learning robust visual features without supervision.
\newblock \emph{arXiv preprint arXiv:2304.07193}, 2023.

\bibitem[Pavlakos et~al.(2019)Pavlakos, Choutas, Ghorbani, Bolkart, Osman, Tzionas, and Black]{pavlakos2019expressive}
Georgios Pavlakos, Vasileios Choutas, Nima Ghorbani, Timo Bolkart, Ahmed~AA Osman, Dimitrios Tzionas, and Michael~J Black.
\newblock Expressive body capture: 3d hands, face, and body from a single image.
\newblock In \emph{Proceedings of the IEEE/CVF conference on computer vision and pattern recognition}, pages 10975--10985, 2019.

\bibitem[Pavlakos et~al.(2024)Pavlakos, Shan, Radosavovic, Kanazawa, Fouhey, and Malik]{pavlakos2024reconstructing}
Georgios Pavlakos, Dandan Shan, Ilija Radosavovic, Angjoo Kanazawa, David Fouhey, and Jitendra Malik.
\newblock Reconstructing hands in 3d with transformers.
\newblock In \emph{Proceedings of the IEEE/CVF Conference on Computer Vision and Pattern Recognition}, pages 9826--9836, 2024.

\bibitem[Peng et~al.(2024)Peng, Zhang, Guo, Cao, and Hu]{peng2024charactergen}
Hao-Yang Peng, Jia-Peng Zhang, Meng-Hao Guo, Yan-Pei Cao, and Shi-Min Hu.
\newblock Charactergen: Efficient 3d character generation from single images with multi-view pose canonicalization.
\newblock \emph{ACM Transactions on Graphics (TOG)}, 2024.

\bibitem[Peng et~al.(2021)Peng, Zhang, Xu, Wang, Shuai, Bao, and Zhou]{peng2021neural}
Sida Peng, Yuanqing Zhang, Yinghao Xu, Qianqian Wang, Qing Shuai, Hujun Bao, and Xiaowei Zhou.
\newblock Neural body: Implicit neural representations with structured latent codes for novel view synthesis of dynamic humans.
\newblock In \emph{Proceedings of the IEEE/CVF Conference on Computer Vision and Pattern Recognition}, pages 9054--9063, 2021.

\bibitem[Podell et~al.(2023)Podell, English, Lacey, Blattmann, Dockhorn, M{\"u}ller, Penna, and Rombach]{podell2023sdxl}
Dustin Podell, Zion English, Kyle Lacey, Andreas Blattmann, Tim Dockhorn, Jonas M{\"u}ller, Joe Penna, and Robin Rombach.
\newblock Sdxl: Improving latent diffusion models for high-resolution image synthesis.
\newblock \emph{arXiv preprint arXiv:2307.01952}, 2023.

\bibitem[Rombach et~al.(2022)Rombach, Blattmann, Lorenz, Esser, and Ommer]{rombach2022highresolutionimagesynthesislatent}
Robin Rombach, Andreas Blattmann, Dominik Lorenz, Patrick Esser, and Björn Ommer.
\newblock High-resolution image synthesis with latent diffusion models, 2022.

\bibitem[Saito et~al.(2019)Saito, Huang, Natsume, Morishima, Kanazawa, and Li]{saito2019pifu}
Shunsuke Saito, Zeng Huang, Ryota Natsume, Shigeo Morishima, Angjoo Kanazawa, and Hao Li.
\newblock Pifu: Pixel-aligned implicit function for high-resolution clothed human digitization.
\newblock In \emph{Proceedings of the IEEE/CVF international conference on computer vision}, pages 2304--2314, 2019.

\bibitem[Saito et~al.(2020)Saito, Simon, Saragih, and Joo]{saito2020pifuhd}
Shunsuke Saito, Tomas Simon, Jason Saragih, and Hanbyul Joo.
\newblock Pifuhd: Multi-level pixel-aligned implicit function for high-resolution 3d human digitization.
\newblock In \emph{Proceedings of the IEEE/CVF conference on computer vision and pattern recognition}, pages 84--93, 2020.

\bibitem[S{\'a}r{\'a}ndi and Pons-Moll(2024)]{sarandi2024neural}
Istv{\'a}n S{\'a}r{\'a}ndi and Gerard Pons-Moll.
\newblock Neural localizer fields for continuous 3d human pose and shape estimation.
\newblock \emph{arXiv preprint arXiv:2407.07532}, 2024.

\bibitem[Shen et~al.(2023)Shen, Guo, Kaufmann, Zarate, Valentin, Song, and Hilliges]{shen2023x}
Kaiyue Shen, Chen Guo, Manuel Kaufmann, Juan~Jose Zarate, Julien Valentin, Jie Song, and Otmar Hilliges.
\newblock X-avatar: Expressive human avatars.
\newblock In \emph{Proceedings of the IEEE/CVF Conference on Computer Vision and Pattern Recognition}, pages 16911--16921, 2023.

\bibitem[Tang et~al.(2023)Tang, Ren, Zhou, Liu, and Zeng]{tang2023dreamgaussian}
Jiaxiang Tang, Jiawei Ren, Hang Zhou, Ziwei Liu, and Gang Zeng.
\newblock Dreamgaussian: Generative gaussian splatting for efficient 3d content creation.
\newblock \emph{arXiv preprint arXiv:2309.16653}, 2023.

\bibitem[Tang et~al.(2025)Tang, Chen, Chen, Wang, Zeng, and Liu]{tang2025lgm}
Jiaxiang Tang, Zhaoxi Chen, Xiaokang Chen, Tengfei Wang, Gang Zeng, and Ziwei Liu.
\newblock Lgm: Large multi-view gaussian model for high-resolution 3d content creation.
\newblock In \emph{European Conference on Computer Vision}, pages 1--18. Springer, 2025.

\bibitem[Voleti et~al.(2025)Voleti, Yao, Boss, Letts, Pankratz, Tochilkin, Laforte, Rombach, and Jampani]{voleti2025sv3d}
Vikram Voleti, Chun-Han Yao, Mark Boss, Adam Letts, David Pankratz, Dmitry Tochilkin, Christian Laforte, Robin Rombach, and Varun Jampani.
\newblock Sv3d: Novel multi-view synthesis and 3d generation from a single image using latent video diffusion.
\newblock In \emph{European Conference on Computer Vision}, pages 439--457. Springer, 2025.

\bibitem[Wang et~al.(2024)Wang, Jiang, Xu, Zhang, Wang, Zhang, Cao, Cao, Wang, and Fu]{wang2024vividpose}
Qilin Wang, Zhengkai Jiang, Chengming Xu, Jiangning Zhang, Yabiao Wang, Xinyi Zhang, Yun Cao, Weijian Cao, Chengjie Wang, and Yanwei Fu.
\newblock Vividpose: Advancing stable video diffusion for realistic human image animation.
\newblock \emph{arXiv preprint arXiv:2405.18156}, 2024.

\bibitem[Weng et~al.(2024)Weng, Liu, Tan, Xu, Zhou, Yeung-Levy, and Yang]{HumanLRM2024}
Zhenzhen Weng, Jingyuan Liu, Hao Tan, Zhan Xu, Yang Zhou, Serena Yeung-Levy, and Jimei Yang.
\newblock Template-free single-view 3d human digitalization with diffusion-guided lrm.
\newblock \emph{arXiv preprint arXiv:2401.12175}, 2024.

\bibitem[Wu et~al.(2024)Wu, Liu, Cai, Yan, Wang, Hu, Duan, and Ma]{wu2024unique3d}
Kailu Wu, Fangfu Liu, Zhihan Cai, Runjie Yan, Hanyang Wang, Yating Hu, Yueqi Duan, and Kaisheng Ma.
\newblock Unique3d: High-quality and efficient 3d mesh generation from a single image.
\newblock \emph{arXiv preprint arXiv:2405.20343}, 2024.

\bibitem[Wu et~al.(2023)Wu, Zhu, Huang, Zhuang, Lu, and Cao]{wu2023high}
Menghua Wu, Hao Zhu, Linjia Huang, Yiyu Zhuang, Yuanxun Lu, and Xun Cao.
\newblock High-fidelity 3d face generation from natural language descriptions.
\newblock In \emph{Proceedings of the IEEE/CVF Conference on Computer Vision and Pattern Recognition}, pages 4521--4530, 2023.

\bibitem[Xiang et~al.(2024)Xiang, Lv, Xu, Deng, Wang, Zhang, Chen, Tong, and Yang]{xiang2024structured}
Jianfeng Xiang, Zelong Lv, Sicheng Xu, Yu Deng, Ruicheng Wang, Bowen Zhang, Dong Chen, Xin Tong, and Jiaolong Yang.
\newblock Structured 3d latents for scalable and versatile 3d generation.
\newblock \emph{arXiv preprint arXiv:2412.01506}, 2024.

\bibitem[Xiong et~al.(2024)Xiong, Li, Liu, Liao, Hu, Zhu, Ning, Qiu, Wang, Wang, et~al.]{xiong2024mvhumannet}
Zhangyang Xiong, Chenghong Li, Kenkun Liu, Hongjie Liao, Jianqiao Hu, Junyi Zhu, Shuliang Ning, Lingteng Qiu, Chongjie Wang, Shijie Wang, et~al.
\newblock Mvhumannet: A large-scale dataset of multi-view daily dressing human captures.
\newblock In \emph{Proceedings of the IEEE/CVF Conference on Computer Vision and Pattern Recognition}, pages 19801--19811, 2024.

\bibitem[Xiu et~al.(2022)Xiu, Yang, Tzionas, and Black]{xiu2022icon}
Yuliang Xiu, Jinlong Yang, Dimitrios Tzionas, and Michael~J Black.
\newblock Icon: Implicit clothed humans obtained from normals.
\newblock In \emph{2022 IEEE/CVF Conference on Computer Vision and Pattern Recognition (CVPR)}, pages 13286--13296. IEEE, 2022.

\bibitem[Xiu et~al.(2023)Xiu, Yang, Cao, Tzionas, and Black]{xiu2022econ}
Yuliang Xiu, Jinlong Yang, Xu Cao, Dimitrios Tzionas, and Michael~J. Black.
\newblock {ECON: Explicit Clothed humans Optimized via Normal integration}.
\newblock In \emph{Proceedings of the IEEE/CVF Conference on Computer Vision and Pattern Recognition (CVPR)}, 2023.

\bibitem[Xu et~al.(2024{\natexlab{a}})Xu, Cheng, Gao, Wang, Gao, and Shan]{xu2024instantmesh}
Jiale Xu, Weihao Cheng, Yiming Gao, Xintao Wang, Shenghua Gao, and Ying Shan.
\newblock Instantmesh: Efficient 3d mesh generation from a single image with sparse-view large reconstruction models.
\newblock \emph{arXiv preprint arXiv:2404.07191}, 2024{\natexlab{a}}.

\bibitem[Xu et~al.(2024{\natexlab{b}})Xu, Zhang, Liew, Yan, Liu, Zhang, Feng, and Shou]{xu2024magicanimate}
Zhongcong Xu, Jianfeng Zhang, Jun~Hao Liew, Hanshu Yan, Jia-Wei Liu, Chenxu Zhang, Jiashi Feng, and Mike~Zheng Shou.
\newblock Magicanimate: Temporally consistent human image animation using diffusion model.
\newblock In \emph{Proceedings of the IEEE/CVF Conference on Computer Vision and Pattern Recognition}, pages 1481--1490, 2024{\natexlab{b}}.

\bibitem[Yu et~al.(2021)Yu, Zheng, Guo, Liu, Dai, and Liu]{yu2021function4d}
Tao Yu, Zerong Zheng, Kaiwen Guo, Pengpeng Liu, Qionghai Dai, and Yebin Liu.
\newblock Function4d: Real-time human volumetric capture from very sparse consumer rgbd sensors.
\newblock In \emph{Proceedings of the IEEE/CVF conference on computer vision and pattern recognition}, pages 5746--5756, 2021.

\bibitem[Yu et~al.(2024)Yu, Yuan, Cao, Gao, Li, Hu, Quan, Shan, and Tian]{yu2024hifi}
Wangbo Yu, Li Yuan, Yan-Pei Cao, Xiangjun Gao, Xiaoyu Li, Wenbo Hu, Long Quan, Ying Shan, and Yonghong Tian.
\newblock Hifi-123: Towards high-fidelity one image to 3d content generation.
\newblock In \emph{European Conference on Computer Vision}, pages 258--274. Springer, 2024.

\bibitem[Yu et~al.(2023)Yu, Xu, Zhang, Liu, Ye, Wu, Yan, Zhu, Xiong, Liang, et~al.]{yu2023mvimgnet}
Xianggang Yu, Mutian Xu, Yidan Zhang, Haolin Liu, Chongjie Ye, Yushuang Wu, Zizheng Yan, Chenming Zhu, Zhangyang Xiong, Tianyou Liang, et~al.
\newblock Mvimgnet: A large-scale dataset of multi-view images.
\newblock In \emph{Proceedings of the IEEE/CVF conference on computer vision and pattern recognition}, pages 9150--9161, 2023.

\bibitem[Yu et~al.(2020)Yu, Yoon, Lee, Venkatesh, Park, Yu, and Park]{yu2020humbi}
Zhixuan Yu, Jae~Shin Yoon, In~Kyu Lee, Prashanth Venkatesh, Jaesik Park, Jihun Yu, and Hyun~Soo Park.
\newblock Humbi: A large multiview dataset of human body expressions.
\newblock In \emph{Proceedings of the IEEE/CVF Conference on Computer Vision and Pattern Recognition}, pages 2990--3000, 2020.

\bibitem[Zhang et~al.(2024{\natexlab{a}})Zhang, Li, Zhang, Cao, Shan, and Liao]{zhang2024humanref}
Jingbo Zhang, Xiaoyu Li, Qi Zhang, Yanpei Cao, Ying Shan, and Jing Liao.
\newblock Humanref: Single image to 3d human generation via reference-guided diffusion.
\newblock In \emph{Proceedings of the IEEE/CVF Conference on Computer Vision and Pattern Recognition}, pages 1844--1854, 2024{\natexlab{a}}.

\bibitem[Zhang et~al.(2018)Zhang, Isola, Efros, Shechtman, and Wang]{zhang2018unreasonable}
Richard Zhang, Phillip Isola, Alexei~A Efros, Eli Shechtman, and Oliver Wang.
\newblock The unreasonable effectiveness of deep features as a perceptual metric.
\newblock In \emph{Proceedings of the IEEE conference on computer vision and pattern recognition}, pages 586--595, 2018.

\bibitem[Zhang et~al.(2024{\natexlab{b}})Zhang, Yan, Liu, Sheng, and Yang]{zhang20243gen}
Weitian Zhang, Yichao Yan, Yunhui Liu, Xingdong Sheng, and Xiaokang Yang.
\newblock E3gen: Efficient, expressive and editable avatars generation.
\newblock In \emph{Proceedings of the 32nd ACM International Conference on Multimedia}, pages 6860--6869, 2024{\natexlab{b}}.

\bibitem[Zhang et~al.(2024{\natexlab{c}})Zhang, Gu, Wang, Wang, Cheng, Zhu, and Zou]{zhang2024mimicmotion}
Yuang Zhang, Jiaxi Gu, Li-Wen Wang, Han Wang, Junqi Cheng, Yuefeng Zhu, and Fangyuan Zou.
\newblock Mimicmotion: High-quality human motion video generation with confidence-aware pose guidance.
\newblock \emph{arXiv preprint arXiv:2406.19680}, 2024{\natexlab{c}}.

\bibitem[Zhang et~al.(2023)Zhang, Sun, Yang, Chen, and Yang]{zhang2023globalcorrelated}
Zechuan Zhang, Li Sun, Zongxin Yang, Ling Chen, and Yi Yang.
\newblock Global-correlated 3d-decoupling transformer for clothed avatar reconstruction.
\newblock In \emph{Advances in Neural Information Processing Systems (NeurIPS)}, 2023.

\bibitem[Zhang et~al.(2024{\natexlab{d}})Zhang, Sun, Yang, Chen, and Yang]{zhang2024global}
Zechuan Zhang, Li Sun, Zongxin Yang, Ling Chen, and Yi Yang.
\newblock Global-correlated 3d-decoupling transformer for clothed avatar reconstruction.
\newblock \emph{Advances in Neural Information Processing Systems}, 36, 2024{\natexlab{d}}.

\bibitem[Zhang et~al.(2024{\natexlab{e}})Zhang, Yang, and Yang]{zhang2024sifu}
Zechuan Zhang, Zongxin Yang, and Yi Yang.
\newblock Sifu: Side-view conditioned implicit function for real-world usable clothed human reconstruction.
\newblock In \emph{Proceedings of the IEEE/CVF Conference on Computer Vision and Pattern Recognition}, pages 9936--9947, 2024{\natexlab{e}}.

\bibitem[Zheng et~al.(2019)Zheng, Yu, Wei, Dai, and Liu]{zheng2019deephuman}
Zerong Zheng, Tao Yu, Yixuan Wei, Qionghai Dai, and Yebin Liu.
\newblock Deephuman: 3d human reconstruction from a single image.
\newblock In \emph{Proceedings of the IEEE/CVF International Conference on Computer Vision}, pages 7739--7749, 2019.

\bibitem[Zheng et~al.(2021)Zheng, Yu, Liu, and Dai]{zheng2021pamir}
Zerong Zheng, Tao Yu, Yebin Liu, and Qionghai Dai.
\newblock Pamir: Parametric model-conditioned implicit representation for image-based human reconstruction.
\newblock \emph{IEEE transactions on pattern analysis and machine intelligence}, 44\penalty0 (6):\penalty0 3170--3184, 2021.

\bibitem[Zhou et~al.(2023)Zhou, Li, Chan, and Loy]{zhou2023propainter}
Shangchen Zhou, Chongyi Li, Kelvin~CK Chan, and Chen~Change Loy.
\newblock Propainter: Improving propagation and transformer for video inpainting.
\newblock In \emph{Proceedings of the IEEE/CVF International Conference on Computer Vision}, pages 10477--10486, 2023.

\bibitem[Zhu et~al.(2024)Zhu, Chen, Dai, Xu, Cao, Yao, Zhu, and Zhu]{zhu2024champ}
Shenhao Zhu, Junming~Leo Chen, Zuozhuo Dai, Yinghui Xu, Xun Cao, Yao Yao, Hao Zhu, and Siyu Zhu.
\newblock Champ: Controllable and consistent human image animation with 3d parametric guidance.
\newblock In \emph{European Conference on Computer Vision (ECCV)}, 2024.

\bibitem[Zhuang et~al.(2022)Zhuang, Zhu, Sun, and Cao]{zhuang2022mofanerf}
Yiyu Zhuang, Hao Zhu, Xusen Sun, and Xun Cao.
\newblock Mofanerf: Morphable facial neural radiance field.
\newblock In \emph{European conference on computer vision}, pages 268--285. Springer, 2022.

\bibitem[Zhuang et~al.(2023{\natexlab{a}})Zhuang, Zhang, Feng, Zhu, Yao, Li, Cao, Shan, and Cao]{zhuang2023anti}
Yiyu Zhuang, Qi Zhang, Ying Feng, Hao Zhu, Yao Yao, Xiaoyu Li, Yan-Pei Cao, Ying Shan, and Xun Cao.
\newblock Anti-aliased neural implicit surfaces with encoding level of detail.
\newblock In \emph{SIGGRAPH Asia 2023 Conference Papers}, pages 1--10, 2023{\natexlab{a}}.

\bibitem[Zhuang et~al.(2023{\natexlab{b}})Zhuang, Zhang, Wang, Zhu, Feng, Li, Shan, and Cao]{zhuang2023neai}
Yiyu Zhuang, Qi Zhang, Xuan Wang, Hao Zhu, Ying Feng, Xiaoyu Li, Ying Shan, and Xun Cao.
\newblock Neai: A pre-convoluted representation for plug-and-play neural ambient illumination.
\newblock \emph{arXiv preprint arXiv:2304.08757}, 2023{\natexlab{b}}.

\bibitem[Zhuang et~al.(2024)Zhuang, He, Zhang, Wang, Zhu, Yao, Zhu, Cao, and Zhu]{zhuang2024towards}
Yiyu Zhuang, Yuxiao He, Jiawei Zhang, Yanwen Wang, Jiahe Zhu, Yao Yao, Siyu Zhu, Xun Cao, and Hao Zhu.
\newblock Towards native generative model for 3d head avatar.
\newblock \emph{arXiv preprint arXiv:2410.01226}, 2024.

\end{thebibliography}
}

\clearpage
\setcounter{page}{1}
\appendix
\maketitlesupplementary

In this supplementary material, we provide additional details and visualizations to support the claims made in our main paper. Sec.~\ref{dataset} provides further details on the \datasetname dataset, including visualizations, important statistics, and the methodology to enhance the 3D consistency and diversity of multi-view images.
Sec.~\ref{idol} describes the training procedure and setup for our proposed method, \modelname.
Sec.~\ref{experiment} presents additional experimental results, including comparison tables and results from the user study.

\begin{figure*}[ht]
  \centering

  \includegraphics[width=1\linewidth]{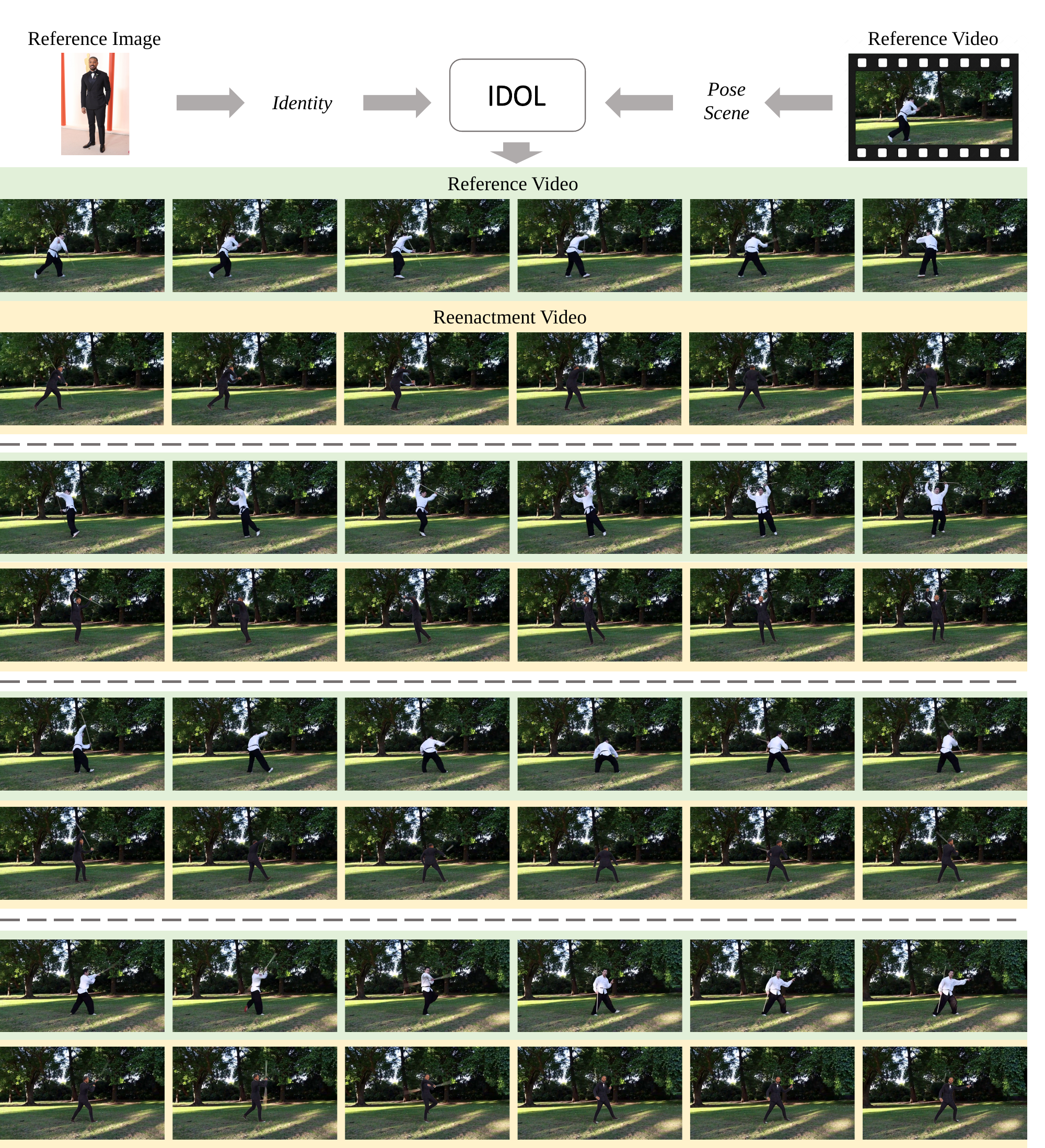}
\caption{The visualization of the reenactment.}
\label{fig:reenact}
\end{figure*}

\section{More Details of \datasetname Data}
\label{dataset}
This section provides a more detailed explanation of the \datasetname dataset generation process, along with additional visualizations.  Sec~\ref{sec2_supp:enhancing_3D_consistency} describes our approach to improving the 3D consistency of MVChamp during training, while Sec~\ref{sec2_supp:inference} presents the prompt template and attribute set used to generate reference images, as well as the generation process with MVChamp. Sec.~\ref{sec2_supp:dataset_add_vis} demonstrates more visualization of the dataset.

\subsection{Improving 3D Consistency of Image Animation}
\label{sec2_supp:enhancing_3D_consistency}

Champ~\cite{zhu2024champ} is one of the state-of-the-art Human Image Animation models, enhanced by multiple conditions rendered from DWPose and SMPL. We employ a two-stage training process to enhance the 3D consistency of the Champ model for human multi-view synthetic, referred to as MVChamp.

\paragraph{Fine-tuning Champ on Large-scale Human Videos with Whole-body Conditions}
To enable Champ to learn more human 3D prior knowledge, we curate a dataset of approximately 100K dance videos for fine-tuning, of which around 20K explicitly contain human turning motions. Full parameter training of MVChamp on such a large dance dataset effectively enhances its understanding of human 3D prior knowledge. Additionally, we employ HaMeR~\cite{pavlakos2024reconstructing}, a state-of-the-art model for 3D hand reconstruction, to specifically reconstruct hand poses from images. These reconstructed hand poses are rendered into depth maps and used as an additional pose control signal for precise whole-body reconstruction and animation.

\paragraph{Fine-Tuning Temporal Blocks on 3D Human Dataset}
We use the open-source scanned dataset THuman 2.1~\cite{zheng2019deephuman}, rendered in Blender, to produce 24 uniformly sampled views along the horizontal dimension to fine-tune the temporal layers of MVChamp using standard diffusion loss.

\paragraph{Improving Temporal Consistency from the First to Last Frames}
Although the MVChamp model generates highly continuous multi-view images between adjacent frames, significant discrepancies remain between the first and last views, even though these two views are continuous in content. This issue likely arises from the model's emphasis during training on ensuring continuity between adjacent frames while neglecting the larger temporal gap between the first and last frames. Thus, we propose the \emph{Temporal Shift Denoising Strategy} to address this issue. During each denoising step, we shift the current latent inputs and pose condition signals along the temporal axis, moving the latent inputs and pose condition of the last frame to the first frame. This strategy ensures that each frame can access contextual information during most of the denoising steps, effectively eliminating discrepancies between the first and last frames at the same inference cost. %

\subsection{Generating Balanced and Diverse Images}
\label{sec2_supp:inference}

Balanced, diverse, high-quality, high-resolution, and full-body images are scarce in existing human-centric datasets, and they are challenging to collect on the Internet due to copyright and portrait rights issues. Therefore, we mix the real-life images and generate photorealistic images to obtain the large-scale quantity and high-quality images. Specifically, we extract approximately 10,000 real-life images from the open-source dataset DeepFashion~\cite{liu2016deepfashion} and use Flux~\cite{flux}, a state-of-the-art text-to-image model, to generate balanced and diverse human reference images. We ensure balance and diversity across five dimensions during image generation: \textit{area}, \textit{clothing}, \textit{body shape}, \textit{age} and \textit{gender}. Each dimension value is randomly selected from a large set of options generated by GPT-4~\cite{achiam2023gpt}, with prompt templates as follows: \textit{
Front view, full-body pose of a \{age\} old \{body shape\} \{area\} \{gender\} wearing \{clothing\} and visible hands. He/She stands against a white background, evenly lit.} Ultimately, we collect a total of 100,000 balanced and diverse full-body human reference images. 

For each dimension, the possible options are as follows:
    \begin{enumerate}
        \item \textbf{Area}: \textit{United States, Canada, Mexico, Guatemala, Cuba, Brazil, Argentina, Colombia, Chile, Peru, United Kingdom, Germany, France, Italy, Spain, Netherlands, Belgium, Switzerland, Poland, Sweden, Nigeria, Egypt, South Africa, Kenya, Morocco, Ghana, Tanzania, Ethiopia, Uganda, Algeria, Saudi Arabia, Iran, Turkey, Israel, United Arab Emirates, Qatar, Kuwait, Jordan, Oman, Lebanon, Kazakhstan, Uzbekistan, Turkmenistan, Kyrgyzstan, Tajikistan, India, Pakistan, Bangladesh, Sri Lanka, Nepal, Bhutan, China, Japan, South Korea, Mongolia, North Korea, Indonesia, Thailand, Vietnam, Malaysia, Philippines, Singapore, Myanmar, Cambodia, Laos, Brunei, Australia, New Zealand, Papua New Guinea, Fiji, Solomon Islands, Jamaica, Haiti, Dominican Republic, Puerto Rico, Trinidad and Tobago, Panama, Costa Rica, Nicaragua, Honduras, El Salvador, Belize, etc.}

        \item \textbf{Clothing}: \textit{T-shirts, Jeans, Casual pants, Dresses, Shorts, Tank tops, Sweaters, Cardigans, Jumpsuits, Hoodies, Suits, Business shirts, Formal skirts, Dress pants, Blazers, Tie, Waistcoats, Formal shoes, Briefcases, Leather belts, Sport shirts, Fitness clothes, Sports shoes, Tracksuits, Gym shorts, Leggings, Swimwear, Cycling gear, Compression wear, Evening gowns, Tuxedos, Long dresses, Tailcoats, Cocktail dresses, Party wear, Ceremonial suits, Ball gowns, Dress shoes, Fine jewelry, Hiking clothes, Waterproof jackets, Thermal wear, Camping gear, Fishing vests, Hunting apparel, Snowboarding pants, Rain boots, Cotton shirts, Linen dresses, Chiffon blouses, Sandals, Sunglasses, Short sleeves, Beachwear, Crop tops, Wool coats, Thick cotton sweaters, Fur jackets, Beanies, Boots, Gloves, Scarves, Thermal leggings, Padded parkas, Insulated boots, Hanfu, Kimono, Sari, African tribal dresses, Scottish kilts, Bavarian lederhosen, Moroccan kaftans, Hawaiian shirts, Russian ushankas, Streetwear, Avant-garde designs, Fusion wear, Boho chic, Minimalist styles, High fashion, Urban outfits, Eco-friendly clothing, Techwear, Nurse uniforms, Firefighter gear, Construction vests, Police uniforms, Military boots, Lab coats, Coveralls, Military uniforms, Academic gowns, Judicial robes, Clerical vestments, Diplomatic suits, Regalia, etc.}

        \item \textbf{Body shape}: \textit{Slight, Lean, Petite, Athletic, Fit, Average, Built, Buff, Bodybuilder, Full-figured, Stocky, Large.}

        \item \textbf{Age}: \textit{20--30 years, 30--40 years, 40--50 years, 50--60 years, 60--70 years, 70--80 years, 80--90 years.}
        \item \textbf{Gender}: \textit{Female and male.}
    \end{enumerate}

\subsection{Additional Visualization}
\label{sec2_supp:dataset_add_vis}

Fig.~\ref{sup:flux_images} shows the diversity of reference images generated using our prompt template and attribute set. Fig.\ref{supp:dataset_vis} and Fig.~\ref{supp:dataset_vis_2} illustrate the multi-view images under diverse poses generated by our MVChamp.

\subsection{Application: Human Video Reenactment}
The goal of this application is to replace a person in a reference video with a new identity while preserving the background and pose. Given a reference image that provides the target identity, and a reference video that provides the pose and background of the original person, the task is to seamlessly swap the person in the video while maintaining the integrity of the scene. We visualize the results in Fig. \ref{fig:reenact}.

To achieve this, we follow a multi-step process:

\textbf{Identity Reconstruction}: The \modelname model is used to reconstruct an animatable 3D human from the reference image. This model generates a highly detailed and realistic representation of the target identity, allowing us to manipulate the avatar to match various poses.

\textbf{Background Inpainting}: The video inpainting process restores the regions of the video frame where the original person has been replaced, ensuring a seamless background. It involves detecting and tracking the target area using a segmentation method, which is initialized and refined by the widely used zero-shot segmentation model, Segment Anything Model (SAM)\cite{kirillov2023segment}. Once the target area is segmented and tracked, the remaining regions are completed using the video inpainting method, ProPainter\cite{zhou2023propainter}, ensuring the background is seamlessly restored with no traces of the replaced identity.

\textbf{Pose Animation}: The target pose is extracted from the reference video~\cite{lin2023motion,cai2024smpler}, and the reconstructed human model is animated to match this pose. The IDOL model provides precise control over the 3D human's pose, including fine details such as \textbf{finger movements}, allowing it to adapt dynamically to the reference video’s actions. After animating the 3D human, we render it into the target view and seamlessly blend it with the background.

Utilizing \modelname, our process offers an efficient and high-quality solution for identity replacement in videos, providing greater stability and lower computational cost compared to 2D-based approaches \cite{hu2024animate, zhu2024champ}. This opens up new possibilities for digital content creation and interactive media applications.

\subsection{Representation Comparisons}
\begin{figure}[ht]
  \centering
  \includegraphics[width=1\linewidth]{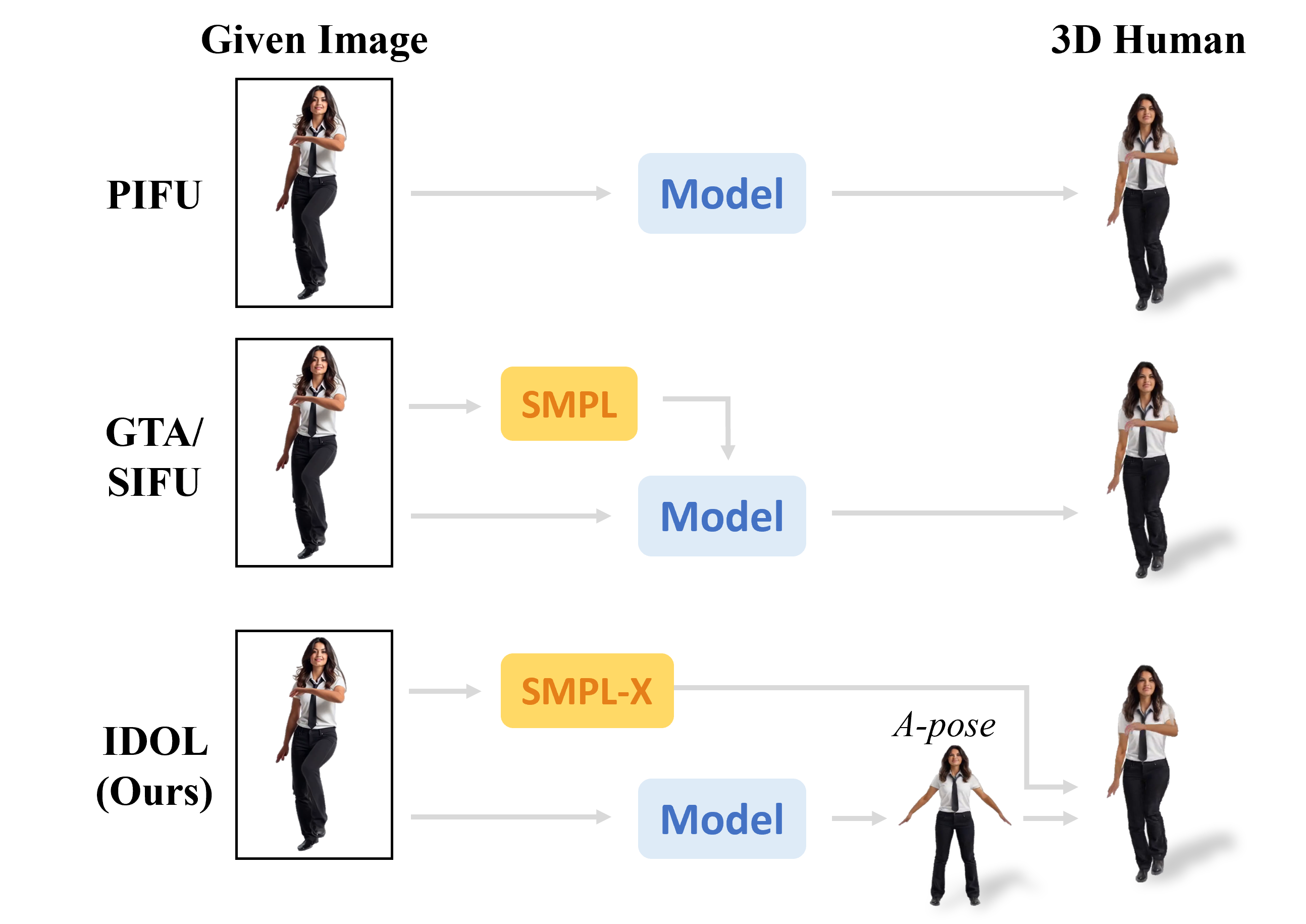}
\caption{Visualization of different approaches for 3D human reconstruction. Unlike PIFu, which directly predicts the 3D human without a parametric prior, and GTA/SIFU, which relies on computationally expensive loop optimization for SMPL alignment, our IDOL method leverages SMPL-X as a prior. This enables more robust and accurate reconstruction while avoiding the pitfalls of error accumulation. Furthermore, our method supports direct animation and editing, enabling additional applications in digital content creation.}
\label{fig:sub_represent}
\end{figure}

\begin{figure*}[ht]
  \centering

  \includegraphics[width=0.85\linewidth]{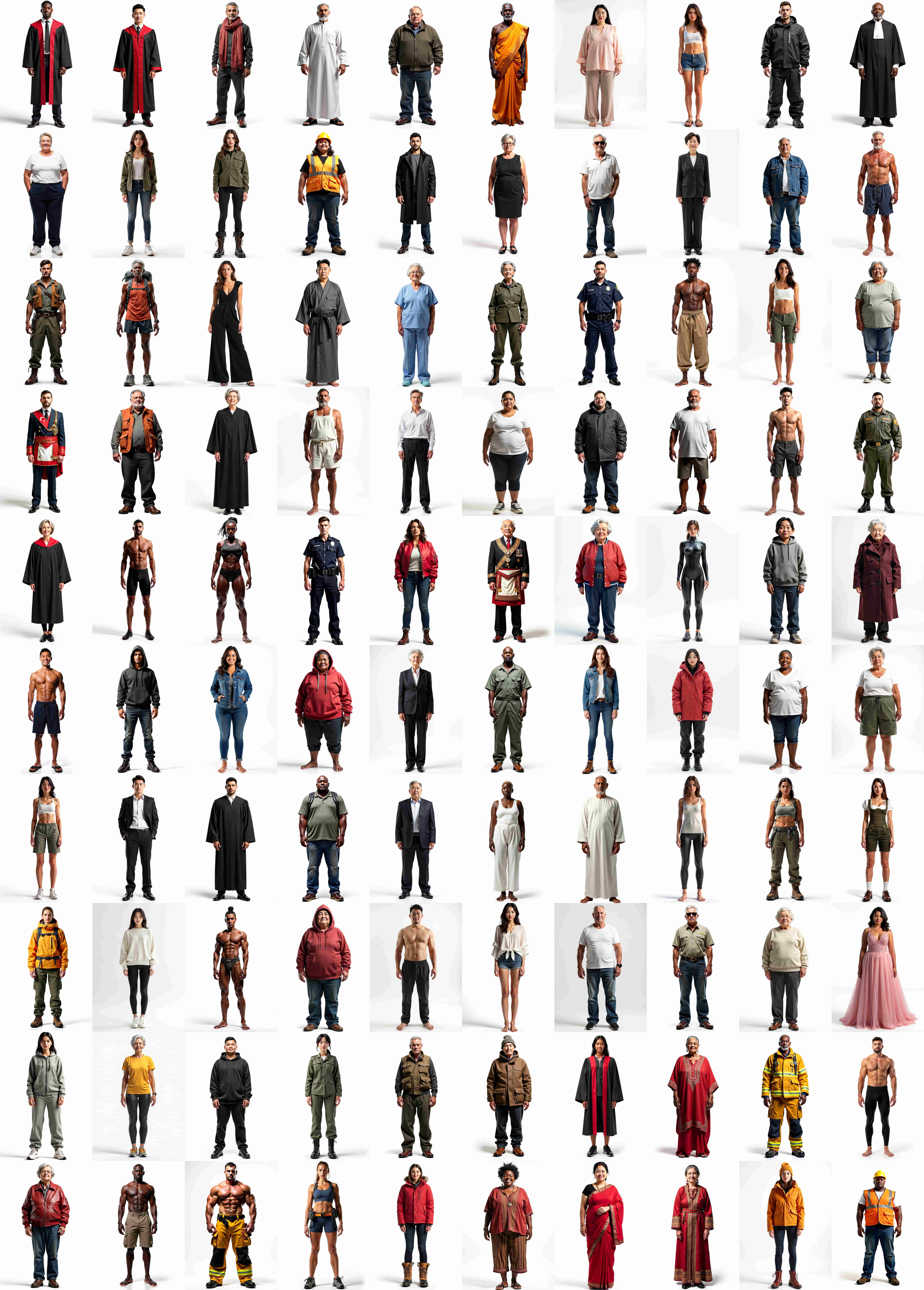}

\caption{Visualization of diverse images generated by Flux~\cite{flux}.}
\label{sup:flux_images}
\end{figure*}

\begin{figure*}[ht]
  \centering

  \includegraphics[width=0.925\linewidth]{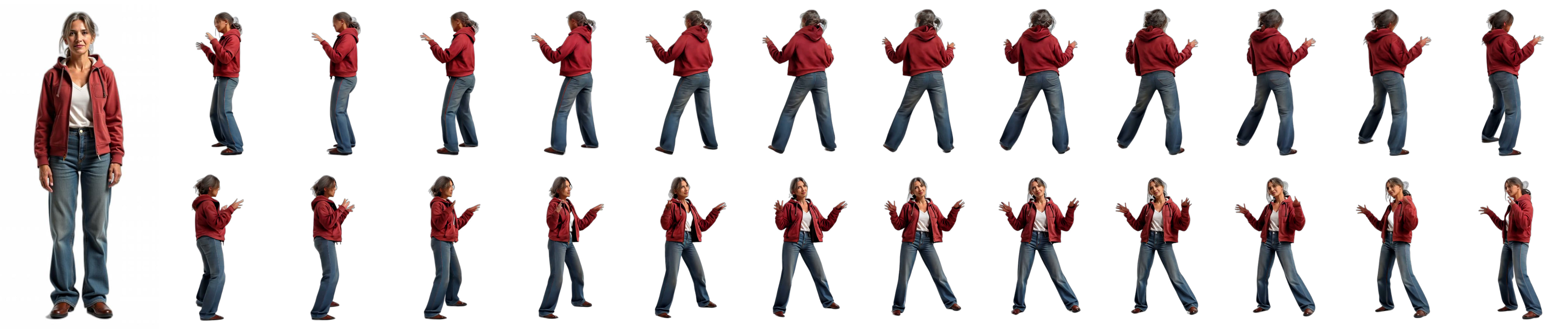}

  \includegraphics[width=0.925\linewidth]{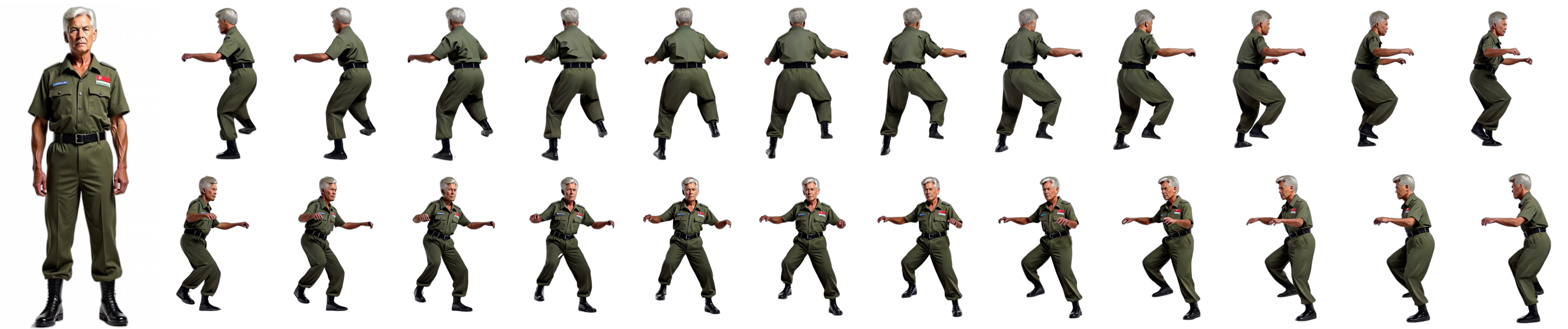}

  \includegraphics[width=0.925\linewidth]{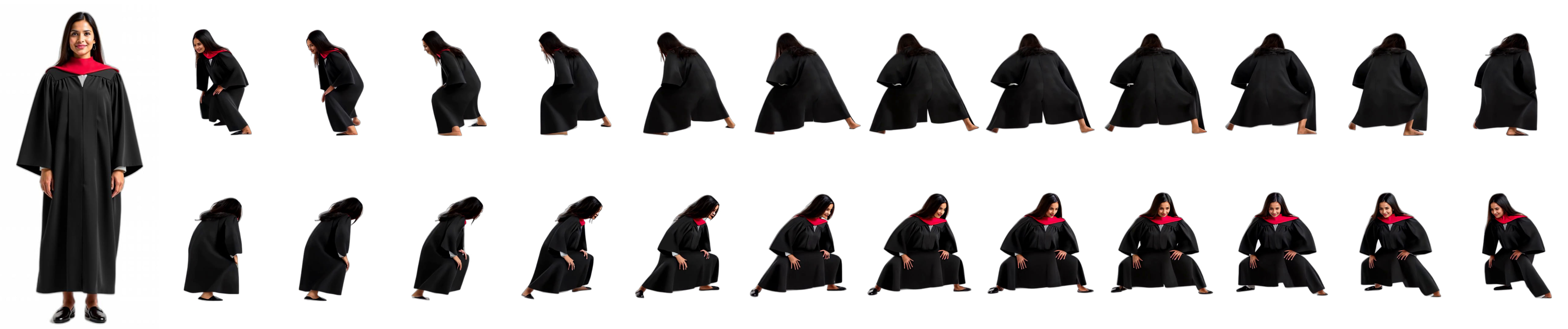}

  \includegraphics[width=0.925\linewidth]{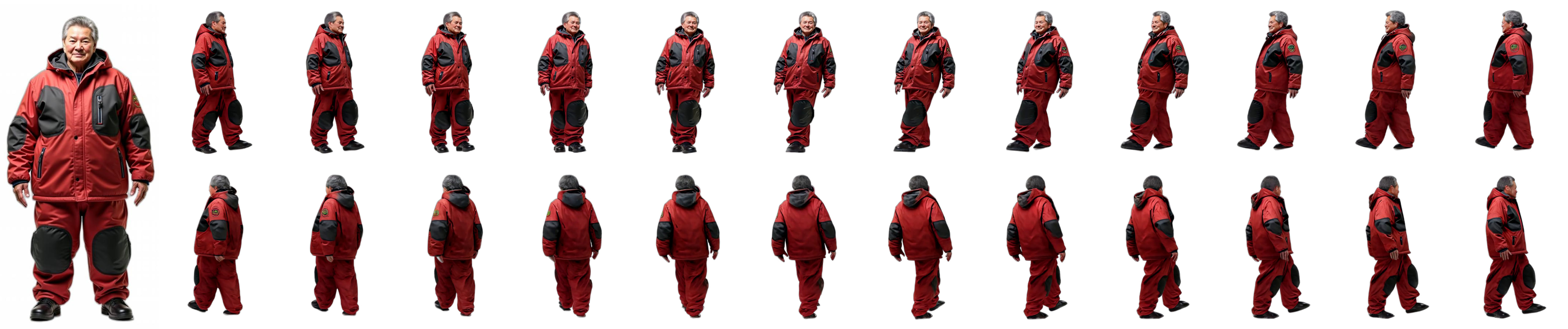}

  \includegraphics[width=0.925\linewidth]{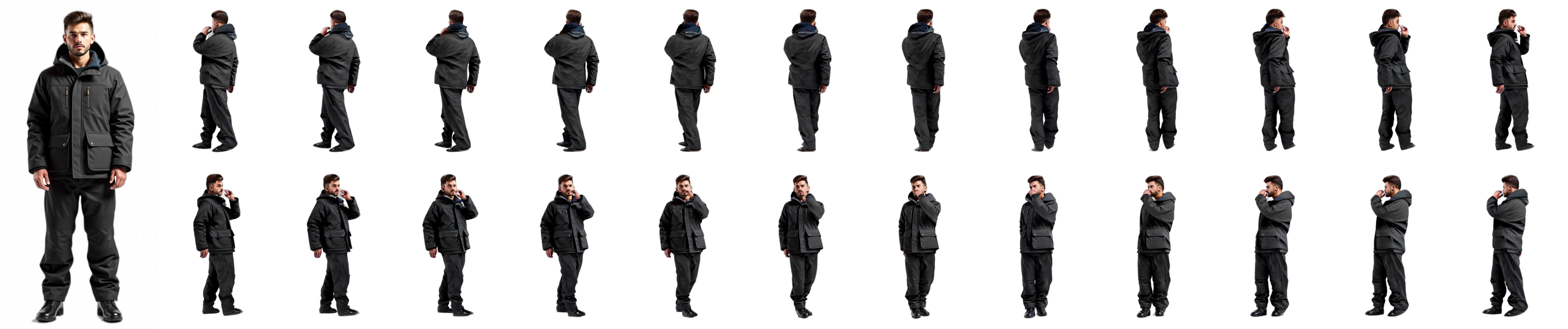}

  \includegraphics[width=0.925\linewidth]{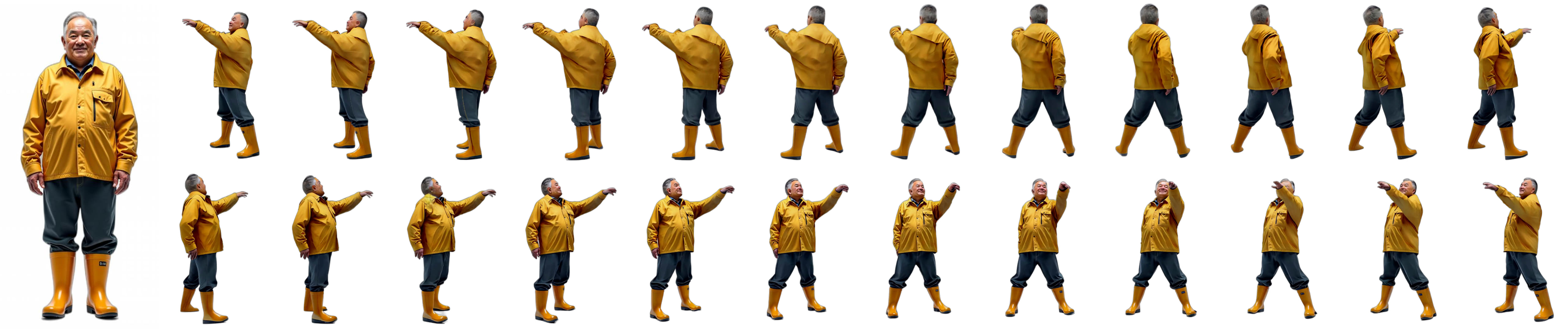}

\caption{Visualization of examples from \datasetname, where the images are generated by Flux and used to generate multi-view images.}
\label{supp:dataset_vis}
\end{figure*}

\begin{figure*}[ht]
  \centering

  \includegraphics[width=0.925\linewidth]{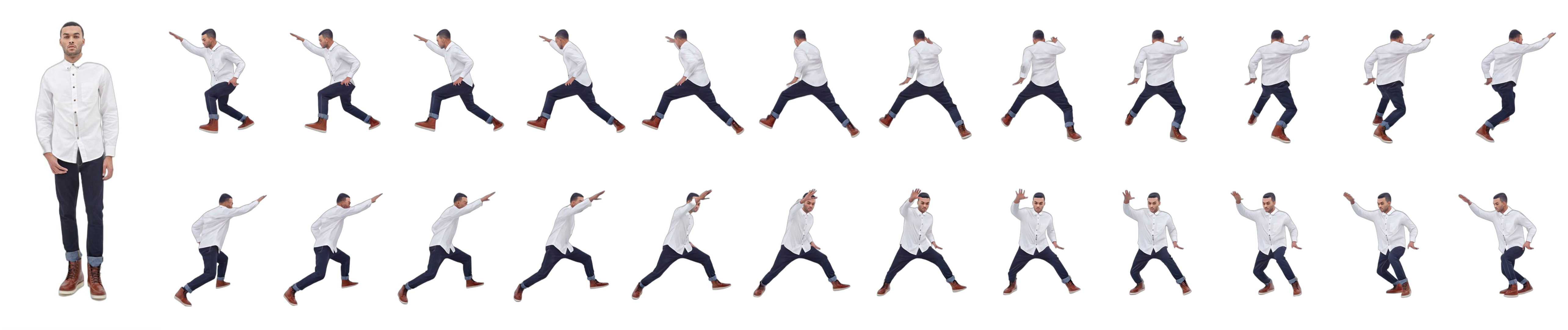}

  \includegraphics[width=0.925\linewidth]{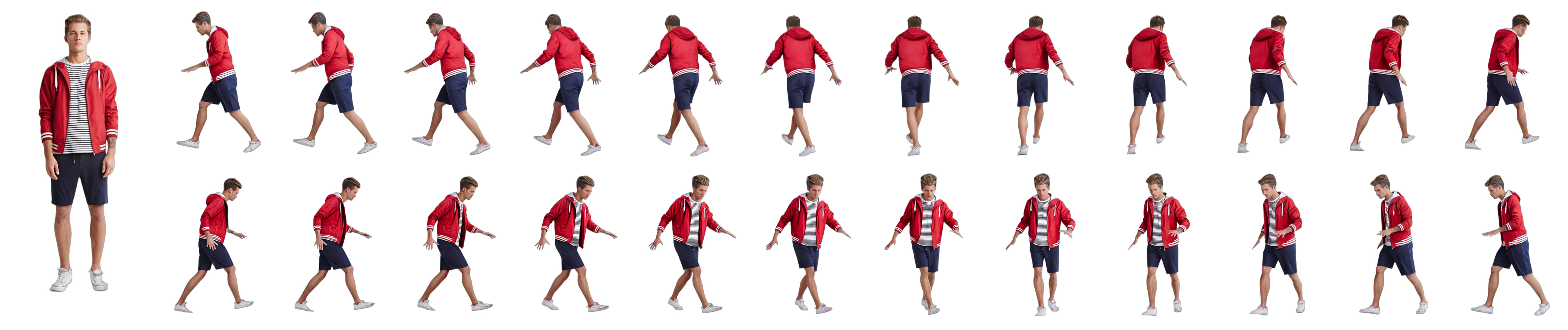}

  \includegraphics[width=0.925\linewidth]{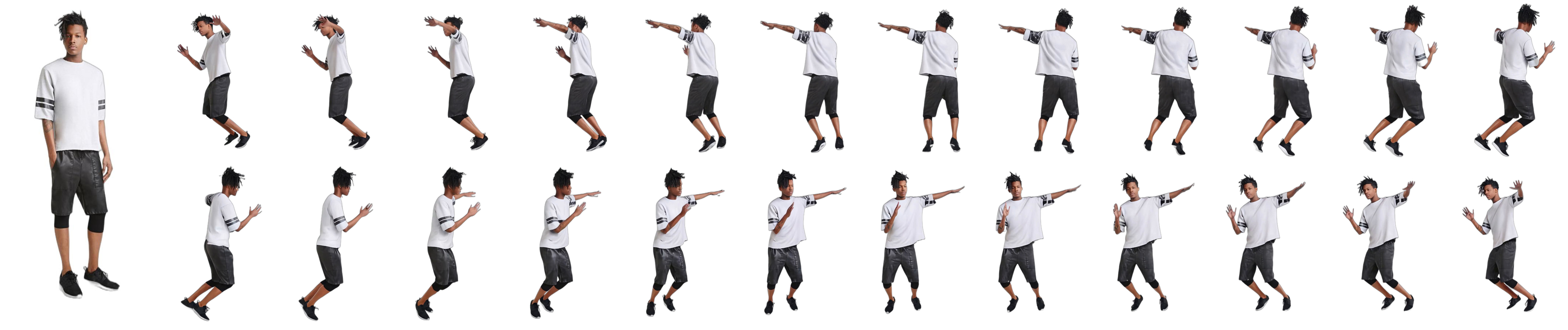}

  \includegraphics[width=0.925\linewidth]{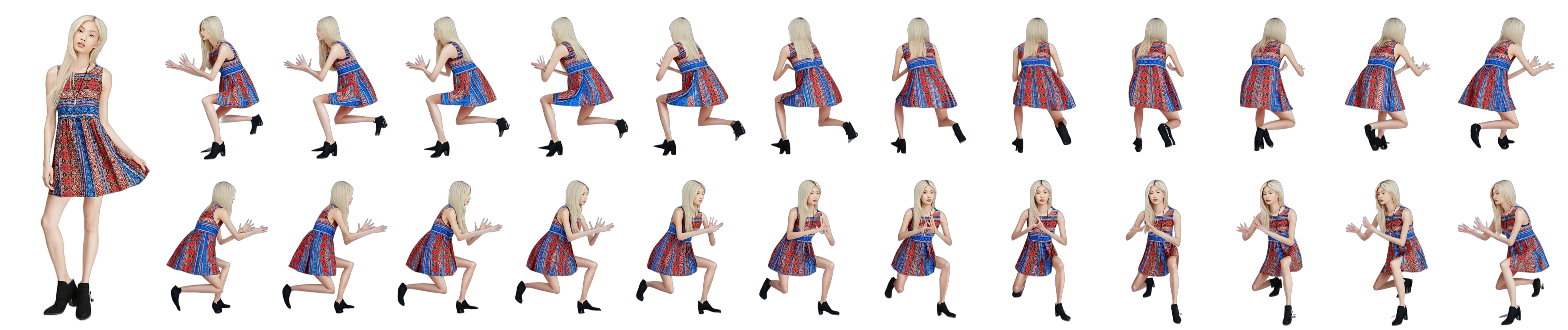}

  \includegraphics[width=0.925\linewidth]{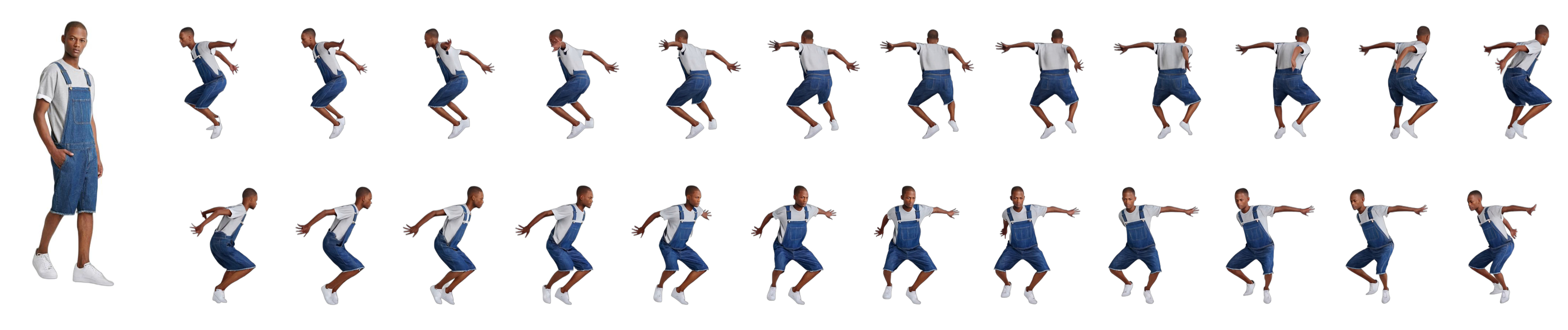}

  \includegraphics[width=0.925\linewidth]{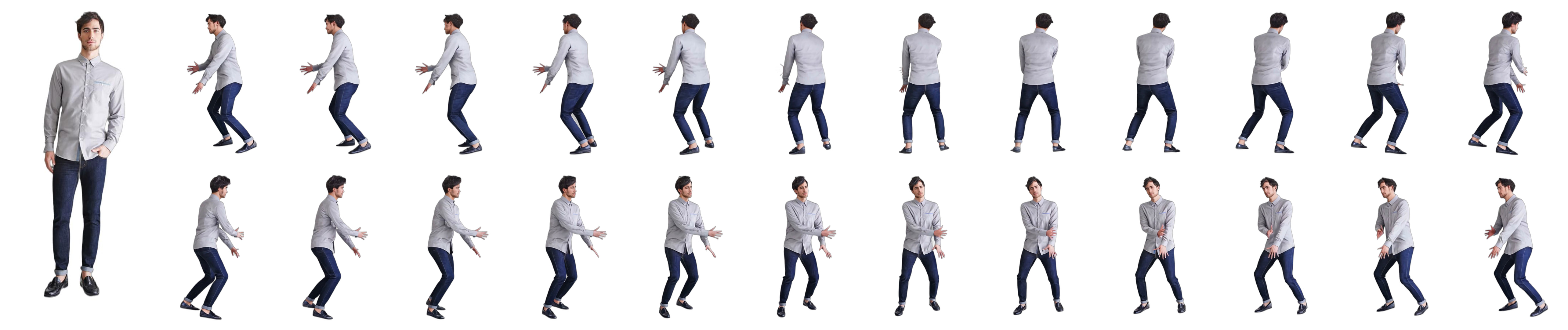}

\caption{Visualization of examples from \datasetname, where the images are derived from the DeepFashion~\cite{liu2016deepfashion} dataset and used to generate multi-view images.}
\label{supp:dataset_vis_2}
\end{figure*}

To further illustrate the differences between our method and previous approaches, we provide a comparison in Fig.~\ref{fig:sub_represent}.
Below, we explain the key differences:

\textbf{Comparison to PIFU}:
PIFu predicts the 3D human shape directly from a given image without leveraging a parametric model prior. While effective for simple cases, it often lacks robustness and precision, particularly when handling challenging poses or incomplete observations~\cite{xiu2022icon}.

\textbf{Comparison to GTA/SIFU}:
GTA and SIFU utilize loop optimization~\cite{xiu2022icon, xiu2022econ} to align the reconstructed output with SMPL models. While this alignment step is crucial for pixel-aligned operations~\cite{saito2019pifu}, it introduces several significant drawbacks:

- High computational cost: Loop optimization requires multiple iterations, adding several minutes of processing time. Additionally, it depends on the estimation of intermediate representations such as masks, normals, and skeletons.

- Error accumulation: Misalignments during optimization can accumulate over iterations, degrading the quality of the final 3D human reconstruction.

\textbf{Our Approach}:
In contrast, our method adopts a direct and efficient pipeline:
We extract image features using a large-scale encoder~\cite{khirodkar2025sapiens}, which captures rich and detailed visual information.
We then predict the 3D human shape and appearance in a uniform space, directly providing the 3D human reconstruction along with the estimated SMPL-X parameters.

By decoupling feature extraction from SMPL-X-based 3D prediction, our approach avoids the error accumulation inherent in optimization-based methods. When pose information is unnecessary, our method relies primarily on body shape estimation, reducing the dependency on precise pose alignment. Furthermore, our method supports direct animation and editing (\emph{e.g.}, shape and texture), unlocking additional applications and expanding its potential value in digital content creation.

\section{More Details of \modelname}
\label{idol}
In this section, we describe the training setup and methodology for our proposed method, \modelname.

\subsection{Implement Details}
Our models are trained on a cluster of $32$ NVIDIA H100 GPUs for approximately 1 day, with a batch size of $32$. The optimization is performed using the Adam optimizer with a learning rate of $5e-4$. A warm-up schedule of $3,000$ steps is employed to stabilize training in the initial stages.

The training loss function is a weighted combination of VGG perceptual loss and Mean Squared Error, balanced with a $1:1$ ratio. This loss formulation ensures both perceptual quality and pixel-wise accuracy.

\subsection{Network Architecture}
The proposed network consists of a multi-stage structure designed for high-dimensional feature extraction and reconstruction tasks. The primary components include the pre-trained encoder, UV-Alignment Transformer, and UV decoder. For the encoder, we utilize the large-scale model Sapiens~\cite{khirodkar2025sapiens} to extract and tokenize human features from the input image.

\noindent \textbf{UV-Alignment Transformer.} The neck module employs a hierarchical design inspired by recent advancements in vision transformer architectures \cite{khirodkar2025sapiens}, featuring a decoder embedding layer with a width of $1536$ and $16$ transformer layers. Each transformer encoder layer consists of the following components:
\begin{enumerate}
    \item A layer normalization operation for input stabilization, enhancing training dynamics, and preventing gradient instability.
    \item A multi-head self-attention mechanism that maps inputs into query, key, and value representations, followed by a linear projection layer to integrate attention outputs. This process is regularized through dropout for improved generalization and further normalized to ensure consistent feature scales.
    \item A feed-forward network (FFN) composed of two dense layers with a GeLU activation function applied between them. The FFN architecture is complemented by intermediate normalization layers to enhance stability and improve optimization convergence.
\end{enumerate}

\noindent \textbf{UV Decoder.} The decoder begins by reshaping tokens into a 2D feature map of $64\times64$ resolution. It employs a hierarchical upsampling and convolutional strategy to progressively refine and synthesize outputs. The upsampling mechanism uses transposed convolutional layers to increase spatial resolution, with each stage incorporating normalization and non-linear activation for stable feature transformations. Specifically:
\begin{enumerate}
    \item Upsampling Blocks: The decoder incorporates multiple transposed convolutional layers, which double the spatial resolution at each stage. Instance normalization and \texttt{SiLU} activations provide stable scaling and enable non-linear feature transformations. 
    \item Convolution Block: Three convolutional layers with output channels $\{128, 128, 32\}$ further process the features, applying instance normalization and activation functions to improve feature quality and representation.
\end{enumerate}

\noindent \textbf{Head Module.} Following \cite{zhang20243gen}, we construct two distinct convolutional networks for decoding geometry and color separately. These networks progressively process feature channels, transitioning from an initial channel size of $32$ to the target parameters $\delta_{\mu_k}, \delta_{\mathbf{s}_k},\delta_{\mathbf{r}_k}$ for geometry and $\mathbf{c}_k$ for color.

\section{Experiment}
\label{experiment}
\begin{figure*}[ht]
  \centering

  \includegraphics[width=1\linewidth]{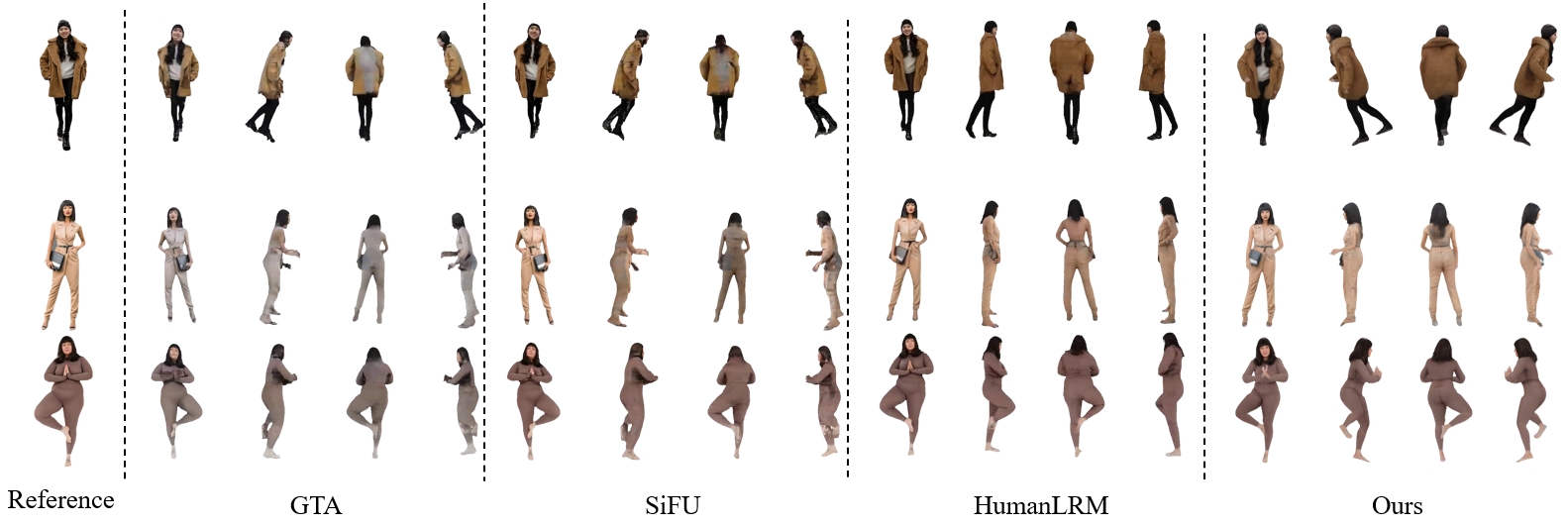}
\caption{More visualization for comparison in the in-the-wild cases. We compare with the reported results by HumanLRM\cite{HumanLRM2024}.}
\label{fig:comp_sup}
\end{figure*}

\begin{figure*}[ht]
  \centering

  \includegraphics[width=0.8\linewidth]{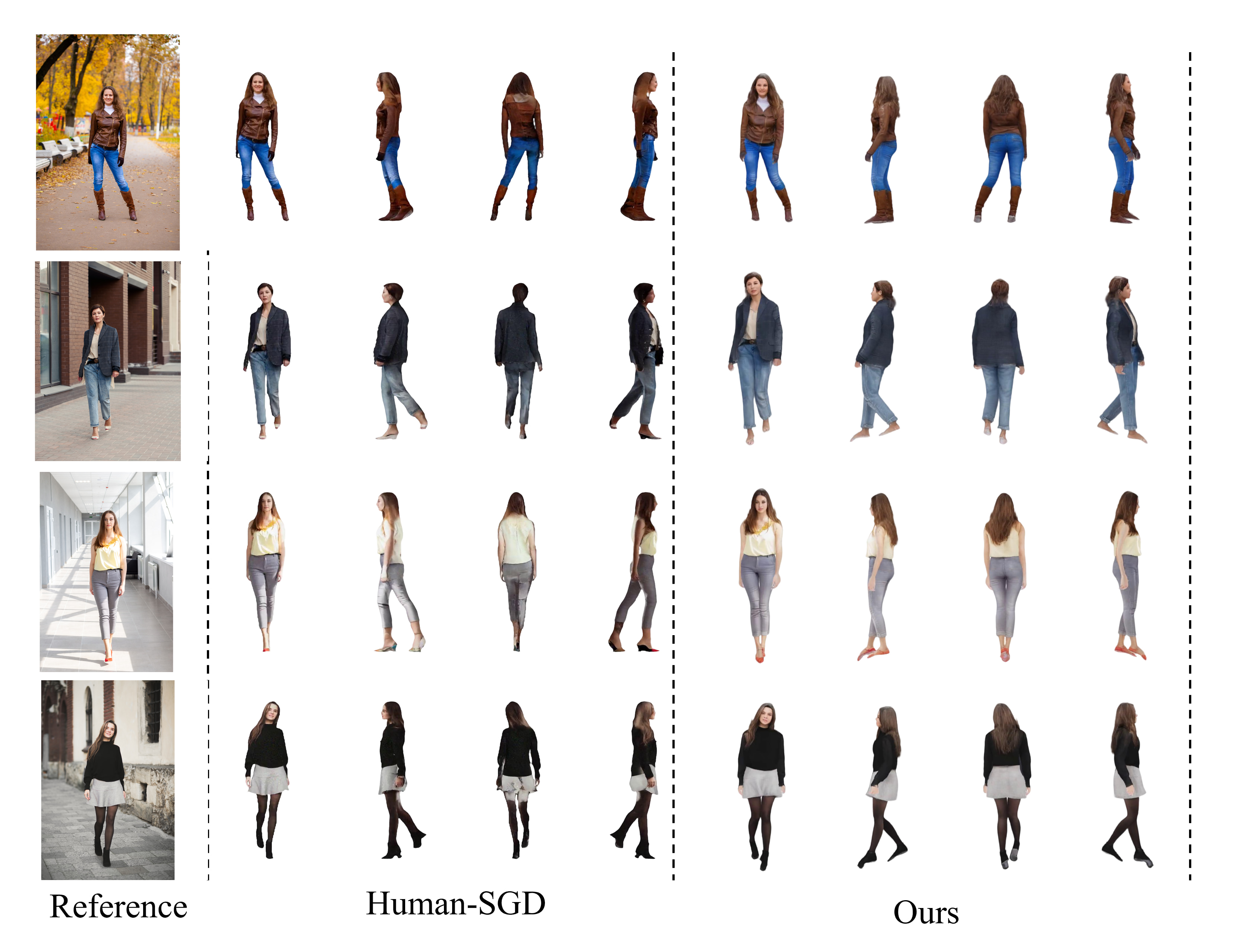}
\caption{More visualization for comparison in the in-the-wild cases. We compare with the reported results by HumanSGD\cite{HumanSGD:2023}.}
\label{fig:comp_sub_sgd}
\end{figure*}

In this section, we present additional experimental results, including comparison tables and the user study. 
We show additional visual comparisons in Fig. \ref{fig:comp_sup} and Fig. \ref{fig:comp_sub_sgd}. We compare with the reported results by \citet{HumanLRM2024} and \citet{HumanSGD:2023}.

\paragraph{More Qualitative Comparisons.} We show additional visual comparisons in Fig. \ref{fig:comp_sup} and Fig. \ref{fig:comp_sub_sgd}. We compare with the reported results by \citet{HumanLRM2024} and \citet{HumanSGD:2023}.

\begin{figure*}[t]

    \centering
   \includegraphics[width=\textwidth]{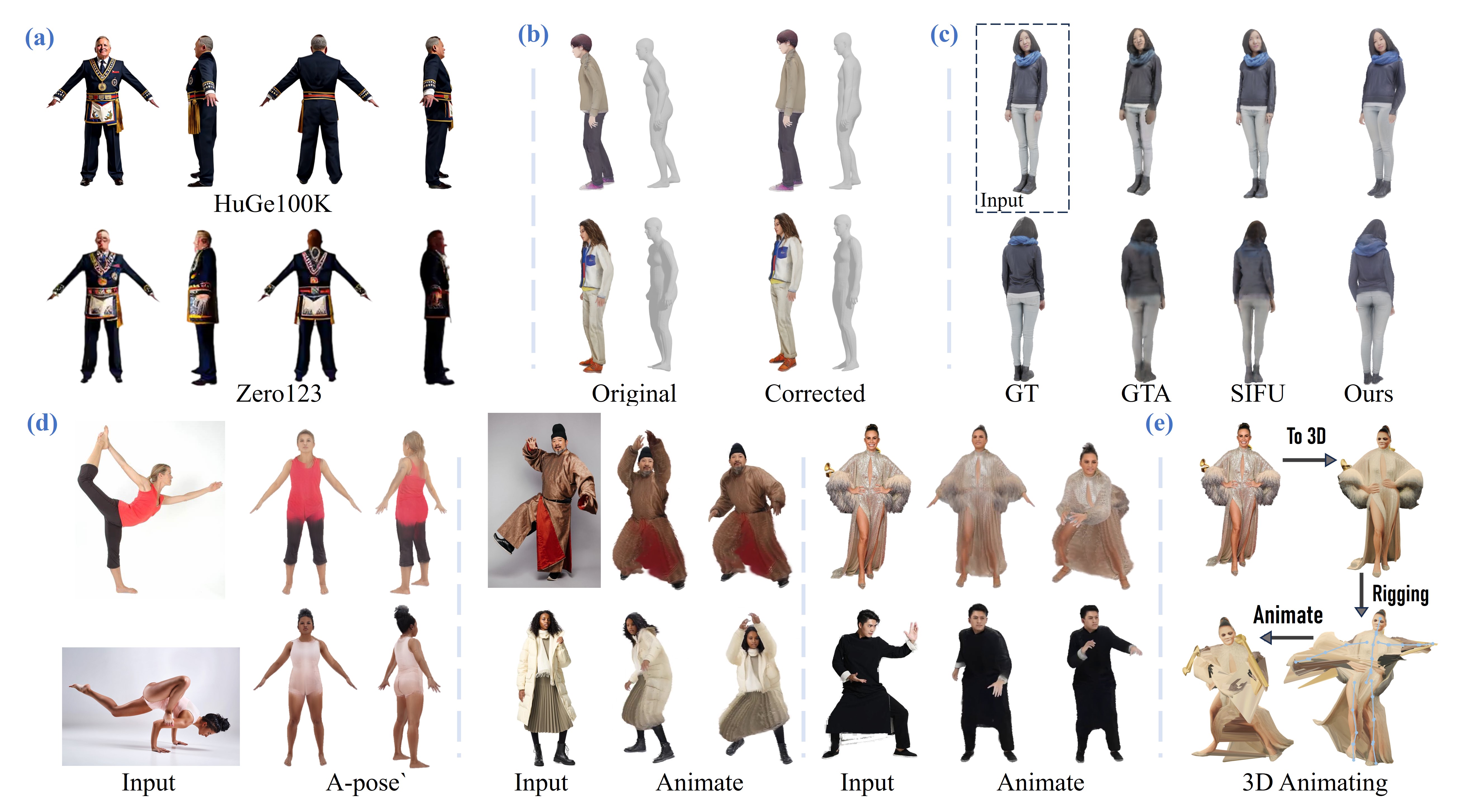}
    \caption{(a) Comparison with Zero123. (b) Original results with leaning/bent poses due to inaccurate SMPL-X, and corrected results with refined SMPL-X. (c) The results on 2K2K. (d) Challenges in large pose and loose cloth animation. (e) 3D animating framework using TRELLIS for image-to-3D and Make-It-Animatable for rigging and animation.}
    \label{fig:res}
    \vspace{-3mm}
\end{figure*}

\paragraph{Effect of the SMPL-X Parameters on Reconstruction.}
Although the reconstruction quality remains good with imperfect SMPL-X input, errors such as leaning or bent shapes can occur due to inaccurate pose parameters, as shown in Fig. 6 of the main content.
This occurs because the avatar is re-posed based on the estimated SMPL-X parameters. Fig.\ref{fig:res}b demonstrates that providing accurate pose information resolves this issue.

\begin{table}[t]
    \centering
    \small
    \begin{tabular}{cc}
        \toprule
        \textbf{Dataset} & \textbf{WE ($\times 10^{-3}$) $\downarrow$} \\
        \midrule
        THuman2.1 & 5.38 \\
        HuGe100K (MVChamp) & 7.33 \\
        Zero123 & 10.51 \\
        \bottomrule
    \end{tabular}
    \caption{\footnotesize Warping Error (WE) comparison across datasets and multi-view synthesis methods, evaluating 3D consistency.}
    \label{tab:warping_error}
\end{table}

\vspace{4mm}

\begin{table}[t]
    \centering
    \small
    \begin{tabular}{cccc}
        \toprule
        \textbf{Method} & \textbf{MSE $\downarrow$} & \textbf{PSNR $\uparrow$} & \textbf{LPIPS $\downarrow$} \\
        \midrule
        SIFU & 0.032 & 15.054 & 1.303 \\
        GTA  & 0.035 & 14.833 & 1.340 \\
        Ours & \textbf{0.023} & \textbf{16.688} & \textbf{1.171} \\
        \bottomrule
    \end{tabular}
    \caption{\footnotesize Quantitative comparison on the 2K2K dataset.}
    \label{tab:comparison_2k2k}
\end{table}

\paragraph{Experimental Comparison with Other Multi-View Image Generation Models.} 
Here, we compare MVChamp with traditional multi-view image generation models based on text-to-image synthesis, specifically Zero123, in the context of human multi-view generation. 
We compare WE~\cite{liu2024evalcrafter} in Tab.~\ref{tab:warping_error}, evaluating 50 random cases. The THuman2.1 dataset serves as the upper bound, and HuGe100K shows comparable results;
Regarding multi-view generation, MVChamp outperforms Zero123 by 30.1\% in 3D consistency. Zero123 generates one novel view at a time, causing multi-view inconsistency. In contrast, MVChamp generates 24 views per batch, ensuring consistency. It also benefits from redundant human priors from dance videos and provides more accurate pose control, enabling well-aligned SMPL-X parameters. See Fig.~\ref{fig:res}a for a visual comparison.

\paragraph{Additional Cases for Evaluating Generalization to Complex Poses and Loose Clothing.}
\label{generalization_exp}
Fig.~\ref{fig:res}d demonstrates IDOL's capability to handle complex poses and loose clothing. This is made possible by our novel architecture, which extracts global features using Sapiens and the diverse HuGe100K dataset. While loose clothing presents challenges due to significant deviation from the body, HuGe100K provides numerous examples, allowing IDOL to recover animatable 3D avatars effectively and reduce issues like tearing in animations, especially in areas such as skirts. \textit{For more examples and animations, please refer to the introduction video (38s-54s).}

\paragraph{Comparison with 3D Animating Methods.}
Classical animation methods typically involve image-to-3D conversion, rigging, and animation. Fig.\ref{fig:res}e demonstrates this pipeline using TRELLIS\cite{xiang2024structured} and Make-It-Animatable~\cite{guo2024make}, which struggles with topology changes, such as detaching the hand from the waist, resulting in artifacts. In contrast, our approach (left) handles these transitions naturally.

\paragraph{Evaluation on Additional 3D Datasets.}
We performed the suggested evaluation on 2K2K using the same settings as in the paper. The quantitative and qualitative results are presented in Tab.\ref{tab:comparison_2k2k} and Fig.\ref{fig:res}c, offering valuable insights into IDOL's generalizability.

\paragraph{User Study.}
\label{experiment}
We conducted a user study with 20 participants via evaluating 50 cases. Participants ranked results based on face, clothing, back-view consistency, and the overall quality. The aggregated results are presented in Tab.~\ref{tab:user_study}, showing the superiority of our method.

\begin{table}[ht]

    \centering
    \begin{tabular}{@{}lcccc@{}}
        \toprule
        Method      & Face & Clothing & Back & Overall \\ 
        \midrule
        GTA        & 0\% & 2.27\% & 0\% & 0\%  \\
        SIFU       & 2.27\% & 4.55\% & 4.55\% & 4.55\%  \\
        HumanLRM   & 45.45\% & 43.18\% & 36.36\% & 45.45\%  \\
        Ours       & 52.28\% & 50.0\% & 59.09\% & 50\%  \\ 
        \bottomrule
    \end{tabular}
    \caption{The user study. We evaluated IDOL on selected cases reported by HumanLRM\cite{HumanLRM2024}, designed to highlight their strengths. Despite the selection for HumanLRM, our method achieves slightly superior performance, demonstrating greater robustness and effectiveness under comparable conditions.}
    \label{tab:user_study}
\end{table}

\end{document}